\documentclass[journal,twoside,web]{ieeecolor}
\usepackage{generic}
\usepackage{cite}
\usepackage{array}
\usepackage{bm}
\usepackage{amsmath,amssymb,amsfonts}
\usepackage{algorithmic}
\usepackage{graphicx}
\usepackage{algorithm,algorithmic}
\usepackage{hyperref}
\usepackage{verbatim}
\usepackage{colortbl}
\usepackage[normalem]{ulem}
\usepackage{booktabs}
\usepackage{multirow}
\usepackage{threeparttable}
\newtheorem{definition}{Definition}
\hypersetup{hidelinks}
\usepackage{textcomp}
\def\BibTeX{{\rm B\kern-.05em{\sc i\kern-.025em b}\kern-.08em
    T\kern-.1667em\lower.7ex\hbox{E}\kern-.125emX}}
\begin{document}

\title{Curvature-Guided Geometric Representation for Protein–Ligand Binding Affinity Prediction}

\author{Shuai Li, Chuan-Xian Ren, Yuhao Li, Ziqi Huang, Yue Pan, Mingzhe Tang and Hong Yan~\IEEEmembership{Fellow,~IEEE}
\thanks{This work is supported in part by National Key R\&D Program of China (2024YFA1011900), National Natural Science Foundation of China (62376291), Guangdong Basic and Applied Basic Research Foundation (2023B1515020004), Science and Technology Program of Guangzhou (2024A04J6413), the Fundamental Research Funds for the Central Universities, Sun Yat-sen University (24xkjc013), and in part by the Hong Kong Innovation and Technology Commission (ITC) (InnoHK Project CIMDA) and the Institute of Digital Medicine of City University of Hong Kong (Project 9229503).}
\thanks{Shuai Li, Chuan-Xian Ren, Yuhao Li, Ziqi Huang, Yue Pan and Mingzhe Tang are with the School of Mathematics, Sun Yat-sen University, Guangzhou 510275, China; Hong Yan is with Department of Electrical Engineering, City University of Hong Kong, 83 Tat Chee Avenue, Kowloon, Hong Kong. Chuan-Xian Ren is the corresponding author (rchuanx@mail.sysu.edu.cn).}}

\maketitle

\begin{abstract}
Protein–ligand binding affinity (PLA) prediction is critical in drug discovery. Despite the notable advancements in machine learning-based approaches, existing methods struggle to jointly characterize local geometric organization and globally coordinated cross-molecular interactions, limiting their ability to model complex binding mechanisms. Here, we propose RicciBind, a geometric representation framework that integrates curvature-guided hierarchical structure learning with optimal transport (OT)-based cross-domain alignment to model molecular interactions. Specifically, RicciBind leverages Ricci curvature to capture local interaction tightness within molecular structures, enhancing structural awareness and organizing atomic interactions into curvature-aware hierarchical representations. An OT-based cluster matching mechanism then aligns protein and ligand clusters across heterogeneous domains under geometric constraints, enabling globally consistent correspondences and revealing higher-order interaction patterns beyond local neighborhoods. By coupling curvature-guided structure encoding with OT-driven cross-domain alignment, RicciBind effectively models complex interaction semantics and substantially improves both the accuracy and interpretability of binding affinity prediction. Extensive experiments demonstrate that RicciBind achieved superior predictive performance and generalization across PLA benchmarks and virtual screening tasks. Ablation studies further confirmed the essential role of Ricci curvature in enhancing molecular interaction representations.
\end{abstract}


\section{Introduction}
\label{sec:introduction}

\IEEEPARstart{P}{redicting} protein–ligand binding affinity (PLA) is a key problem in drug discovery, as PLA quantitatively measures the strength of interaction between a ligand (a small molecule) and a protein (the target). Accurate PLA prediction is critical for identifying promising therapeutic candidates, yet traditional experimental approaches are costly and time-consuming, often requiring years of effort and substantial resources for a single protein–ligand pair~\cite{fleming2018artificial},~\cite{stark2022equibind},~\cite{chen2024multiscale}. Rapid advances in this field have significant potential to enhance the efficiency of drug discovery by facilitating the identification of promising therapeutic candidates.

In recent years, deep learning techniques have attracted considerable attention due to their strong representational capacity and ability to automatically learn relevant features from raw data. By effectively identifying and leveraging task-relevant information, these models enable efficient and accurate predictions, making them a central focus in PLA prediction. To date, deep learning approaches have been developed for predicting PLA~\cite{chen2024multiscale},~\cite{ozturk2018deepdta},~\cite{nguyen2021graphdta},~\cite{yang2023geometric},~\cite{shen2024curvature}. These can be broadly classified into two categories: structure-free models~\cite{ozturk2018deepdta},~\cite{nguyen2021graphdta} and structure-based models~\cite{chen2024multiscale},~\cite{yang2023geometric},~\cite{shen2024curvature}. The fundamental distinction between these approaches lies in whether atomic interactions and three-dimensional (3D) structural information are explicitly incorporated.

\begin{figure}[!hbpt]
    \centering
    \includegraphics[width=1.0\linewidth]{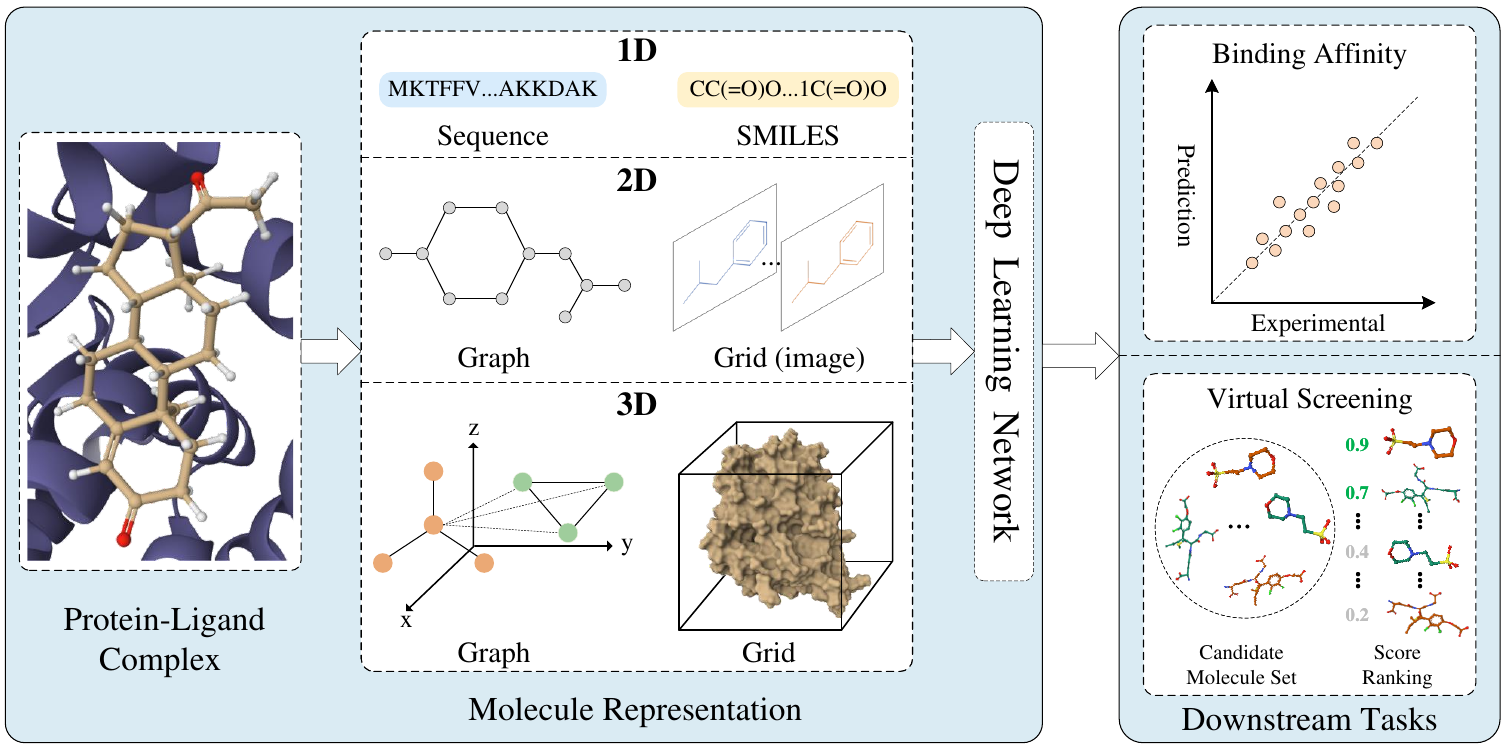}
    \caption{\textbf{Illustration of deep learning paradigms for protein–ligand interaction modeling.} Protein–ligand complexes are represented using structure-free 1D/2D representations or structure-based 3D representations and processed by deep learning networks to learn interaction patterns, supporting downstream tasks such as binding affinity prediction and virtual screening.}
    \label{fig:intro}
\end{figure}

Structure-free methods~\cite{ozturk2018deepdta},~\cite{nguyen2021graphdta},~\cite{torng2019graph},~\cite{li2021co},~\cite{bai2023interpretable} do not rely on 3D structural information of proteins and ligands or explicit interatomic interactions. As illustrated in Fig. \ref{fig:intro}, ligands are typically encoded as simplified molecular-input line-entry system (SMILES) strings, 2D molecular graphs, or 2D grids, while proteins are represented by amino acid sequences. In contrast, structure-based methods~\cite{gomes2017atomic},~\cite{jimenez2018k},~\cite{jiang2021interactiongraphnet},~\cite{feinberg2018potentialnet},~\cite{yang2024interaction},~\cite{zhang2023planet},~\cite{chen2025local} for PLA prediction explicitly incorporate the 3D structures of protein–ligand complexes along with their physicochemical interaction, with 3D graph neural networks (3D-GNNs) emerging as the dominant paradigm. In these approaches, protein–ligand complexes are modeled as spatial interaction graphs, naturally integrating molecular geometry and spatial adjacency relationships to capture informative structural and interaction patterns. Through message-passing schemes defined on relative coordinates or interatomic distances, 3D-GNNs are inherently invariant to global rotations and translations. Owing to these advantages, 3D-GNN-based methods have attracted growing attention for PLA prediction.

Building upon structure-based PLA prediction methods, recent advances have shown that hierarchical representations of 3D molecular graphs, together with multi-level feature extraction, can effectively reduce computational complexity while improving predictive performance~\cite{du2023new},~\cite{kong2024generalist},~\cite{lim2025cheapnet}. Notably, CheapNet~\cite{lim2025cheapnet} captures higher-order molecular representations through differentiable pooling~\cite{ying2018hierarchical}, achieving strong performance in PLA prediction tasks. Despite these advances, fully characterizing molecular interactions still requires an understanding of the intrinsic geometric properties of molecular graphs. However, existing hierarchical methods do not explicitly incorporate such properties, resulting in an incomplete representation of molecular geometry and limiting their ability to capture complex cross-scale dependencies, thereby constraining model expressiveness and interpretability.

To address this limitation, it is crucial to adopt geometric descriptors that can explicitly characterize the intrinsic structure of molecular graphs beyond conventional feature aggregation. Among such descriptors, Ricci curvature, a fundamental concept in differential geometry, provides a principled framework for analyzing and enhancing the graph geometry through its evolution under Ricci flow. It captures both local connectivity patterns and global structural organization of graphs, making it particularly well-suited for graph-based modeling. Its discrete formulations, including Ollivier-Ricci curvature (ORC)~\cite{ollivier2009ricci} and Forman-Ricci curvature (FRC)~\cite{forman2003bochner}, effectively capture structural characteristics of graphs and have been successfully applied across diverse domains such as community detection~\cite{ni2019community},~\cite{sia2019ollivier}, internet topology analysis~\cite{ni2015ricci}, and biomolecular feature representation~\cite{wee2021forman},~\cite{wee2021ollivier}. Notably, ORC has advantages in explaining the over-smoothing and over-squashing phenomena in message-passing graph neural networks~\cite{nguyen2023revisiting}. In recent times, there has been considerable development in the field of curvature-enhanced graph neural network~\cite{ye2019curvature,li2022curvature}. The integration of ORC into graph neural networks has been demonstrated to be a successful strategy in various synthetic and real-world graphs, particularly in large-scale and dense graphs~\cite{shen2024curvature}.

Inspired by the ability of Ricci curvature to capture both local connectivity and global structural organization in graphs, we propose RicciBind, a hierarchical molecular representation framework guided by Ricci curvature that integrates geometric deep learning with optimal transport (OT) theory for protein–ligand interaction modeling and binding affinity prediction. Unlike existing structure-based methods, RicciBind incorporates geometric and topological information encoded by ORC to refine atom-level representations and generate geometrically consistent cluster-level representations. Moreover, RicciBind models proteins and ligands as two heterogeneous structural domains and employs an OT-based cluster matching mechanism to compute bidirectional cross-domain transport plans between protein and ligand cluster distributions, enabling geometrically consistent alignment of key functional clusters while suppressing irrelevant ones, thereby effectively characterizing complex cross-domain interaction patterns.

The key contributions of RicciBind are summarized as follows.
\begin{itemize}
\item A curvature-aware graph embedding module is designed, which takes the protein–ligand interaction graph as input and incorporates ORC as a topological descriptor to enhance atom-level node representations.
\item A curvature-driven clustering module is proposed to adaptively assign atoms to clusters under the guidance of the Ricci curvature, yielding higher-order representations that are more consistent with the geometric–topological structure of molecular graphs.
\item To our knowledge, this is the first study to treat proteins and ligands as two distinct heterogeneous structural domains and to model protein–ligand binding complexes using optimal transport principletheory, offering a theoretical support for the alignment and selection of key functional interaction clusters.
\item RicciBind is comprehensively evaluated across multiple public datasets, with results showing consistently superior performance in protein–ligand affinity prediction and virtual screening compared with existing approaches.
\end{itemize}

The rest of this paper is organized as follows. Section \ref{Related Work} reviews deep learning approaches for predicting PLA and discusses recent advances in the application of Ollivier–Ricci curvature within graph neural networks. Section \ref{Methods} presents the problem formulation and the computation of Ollivier–Ricci curvature, then provides a detailed description of the proposed RicciBind model. Section \ref{Results and Discussion} reports comprehensive experimental results and their analysis. Section \ref{Conclusion} presents a summary and discussion of the study.

\section{Related Work}
\label{Related Work}
\subsection{Deep Learning Methods in PLA Prediction}
The deep learning methods for predicting protein-ligand binding affinity can be categorized into two distinct classifications: structure-based methods and structure-free methods.

Structure-free methods simplify the representation of protein-ligand complexes by neglecting their 3D structural and interaction information. For instance, DeepDTA\cite{ozturk2018deepdta} employs two independent 1D-CNNs to extract features directly from protein and ligand sequences. In contrast, Graph-CNN\cite{torng2019graph} and GraphDTA\cite{nguyen2021graphdta} leverage 2D molecular graphs to improve the feature representation of both proteins and ligands. To enhance protein modeling, DGraphDTA\cite{jiang2020drug} and S-MAN\cite{zhou2020distance} incorporate protein contact maps and spatial distance features. Despite these advancements, such approaches still overlook the molecular intrinsic structures and genuine interaction mechanisms in 3D space, thereby limiting their capacity to attain robust generalization and interpretability.
  
Structure-based methods, by contrast, consider the 3D molecular structural and interaction information between protein and ligand with the aim of improving model generalization and interpretability. These methods can be further categorised into two distinct groups: 3D-CNN based models and 3D-GNN based models. In the 3D-CNN based models, ACNN\cite{gomes2017atomic} develops a general 3D spatial convolution operation for learning atomic-level chemical interactions. $K_{DEEP}$\cite{jimenez2018k} applies a 3D-CNN to voxelized protein and ligand representations enriched with eight pharmacophoric-like features.
However, due to the sparsity of voxelized grid representations and the resulting low computational efficiency of 3D-CNN, recent research has predominantly focused on 3D-GNN based models. At the atomic-level, GIGN\cite{yang2023geometric} and EHIGN\cite{yang2024interaction} represent protein-ligand complexes as heterogeneous graphs that capture both covalent and non-covalent interactions. Hierarchical methods, including LEFTNet\cite{du2023new}, GET\cite{kong2024generalist} and CheapNet\cite{lim2025cheapnet}, extend this framework by integrating block-level and atomic-level information to construct multi-scale representations. These approaches enhance the expressiveness of interaction modeling by capturing protein–ligand relationships across multiple structural hierarchies.

In contrast to existing structure-based methods, RicciBind adopts a hierarchical representation paradigm and explicitly incorporates geometric priors of molecular structures to construct an informative and physically consistent representation of protein–ligand complexes. This design not only enhances the model’s ability to capture complex interaction patterns but also improves the accuracy of binding affinity prediction.

\subsection{Ollivier-Ricci Curvature (ORC) in GNN}

ORC is a discrete analogue of the classical Ricci curvature from differential geometry, which effectively captures the local geometric and topological properties of graphs. Recent studies have revealed a connection between ORC and the issues of ``over-smoothing'' and ``over-squashing'' in message aggregation, demonstrating its effectiveness in alleviating information distortion in message-passing GNNs\cite{toppingunderstanding,nguyen2023revisiting}.
The application of ORC has yielded positive outcomes in a variety of graph neural network models, as evidenced by its notable efficacy in graph learning tasks.  

In the context of the node classification task, the Curvature Graph Network (CurvGN)~\cite{ye2019curvature} integrates ORC into graph convolution network, employing a multi-layer perceptron (MLP) to learn edge curvature for guiding message passing and adapting to local topological structures. Building on this, the Curvature Graph Neural Network (CGNN)\cite{li2022curvature} explicitly converts edge curvature into ORC-related weights through negative curvature processing and curvature normalization modules, further improving the adaptive locality ability of GNNs. In addition, for biomolecular interaction prediction tasks, a few methods have leveraged ORC to capture molecular structures and enrich molecular representations. For instance, Curvature-enhanced Graph Convolutional Network (CGCN)\cite{shen2024curvature} incorporates ORC information of molecular graphs into the node feature aggregation process, employing an MLP to learn edge weight functions from ORC-related vectors.

In this study, RicciBind incorporate ORC into both graph embedding and hierarchical representation stages of molecular modeling, enabling the capture of intrinsic geometric structure and higher-order interaction patterns in protein–ligand complexes. Incorporating ORC enhances the expressive power of protein–ligand complex representations and provides a principled geometric foundation for curvature-guided atomic clustering.

\section{Methods}
\label{Methods}
\begin{figure*}
    \centering
    \includegraphics[width=0.72\linewidth]{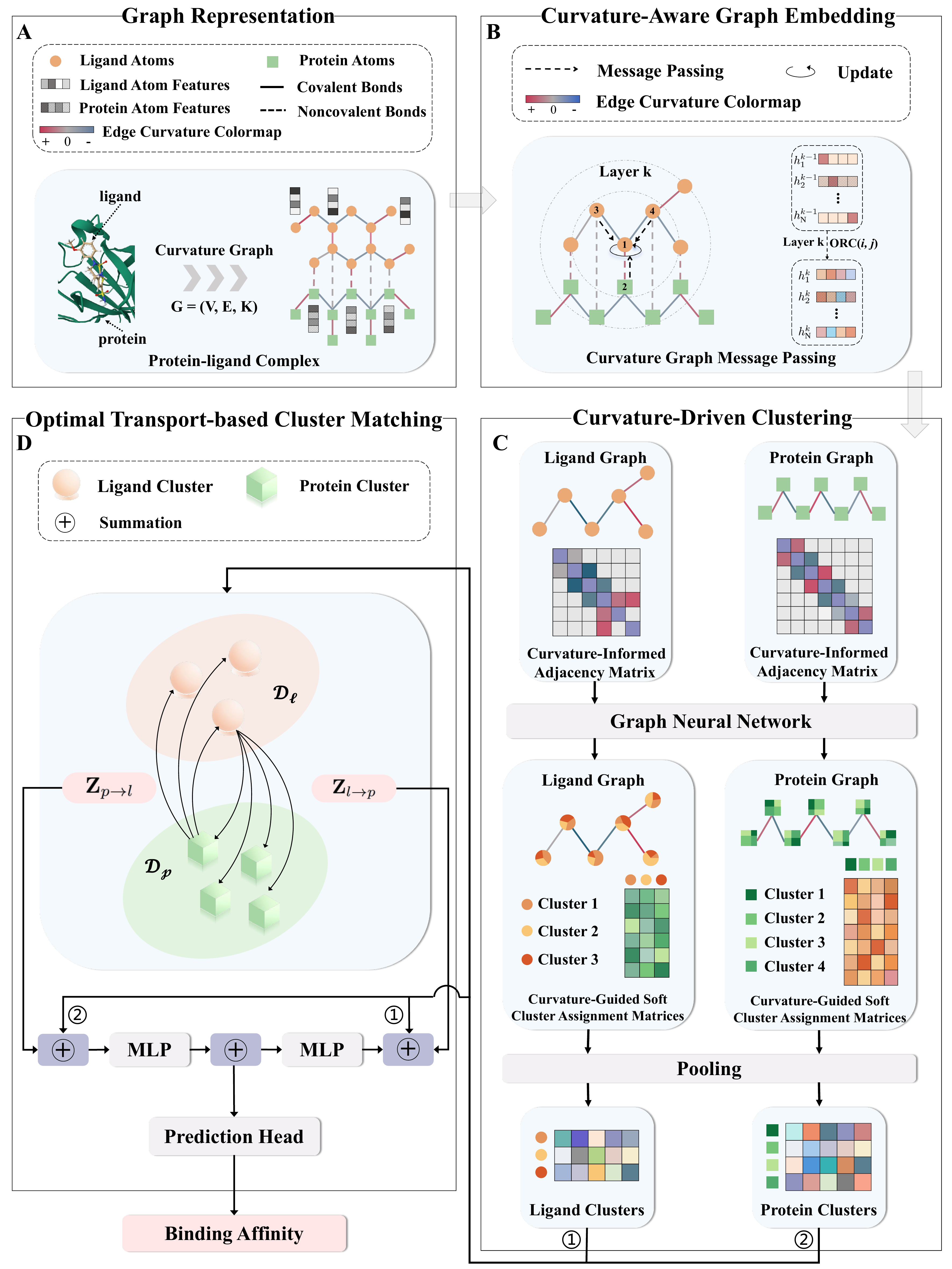}
    \caption{\textbf{Detailed workflow of RicciBind.} \textbf{A}. The graph construction module encodes the protein–ligand complex as a molecular graph. \textbf{B}. The curvature-aware graph embedding module integrates Ricci curvature as a topological descriptor to enhance global atom-level representations. \textbf{C}. The curvature-driven clustering module groups atoms into functional clusters under Ricci curvature guidance. \textbf{D}. The optimal transport-based cluster matching module identifies and aligns functionally critical interaction clusters across domains.}
    \label{fig:framework}
\end{figure*}

This section begins by formalizing the protein–ligand binding affinity prediction problem. It then introduces the definition of Ollivier–Ricci curvature and highlights its geometric significance within molecular graphs. Finally, the RicciBind model is described in detail, with its overall architecture illustrated in Fig.~\ref{fig:framework}.

\subsection{Problem Definition}
In this study, the primary objective is to predict the binding affinity of protein-ligand complexes. Each complex is represented as a 3D molecular graph $G=(V, E, R, K)=(V_l \cup V_p, E_l \cup E_p\cup E_{lp}, R_l \cup R_p,K_l \cup K_p\cup K_{lp})$, where $V_l$ and $V_p$ denote the sets of nodes corresponding to ligand and protein atoms, respectively; $E_l$ and $E_p$ are sets of ligand and protein edges representing covalent bonds, respectively; $E_{lp}$ corresponds to non-covalent interaction edges (i.e., non-covalent bonds) between the ligand and protein; $R_l$ and $R_p$ are the 3D coordinates of ligand and protein nodes; $K_l$ and $K_p$ denote the edge-wise ORC of the ligand and protein; and $K_{lp}$ represents the curvature of edges within the complex. Each node $v_i \in V$ is associated with an initial feature vector $\bm{x}_i \in \mathbb{R}^d$ and a 3D coordinate $\bm{r}_i \in \mathbb{R}^3$. A covalent edge $e_{ij} \in E_l \cup E_p$ exists if a chemical bond is present between nodes $v_i$ and $v_j$, whereas a non-covalent edge $e_{ij} \in E_{lp}$ is defined if the Euclidean distance between a ligand node $v_i \in V_l$ and a protein node $v_k \in V_p$ satisfies $\|\bm{r}_i-\bm{r}_k\|^2 < 5\text{Å}$. The goal is to learn a function $f: G \to \mathbb{R}$ that maps a protein–ligand complex graph $G$ to its predicted binding affinity $\hat{y}=f(G)$.

\subsection{Ollivier-Ricci Curvature}
In the field of Riemannian geometry, Ricci curvature describes several key properties, including the rate of volume growth of geodesic balls, the divergence of geodesics, the transportation distance between balls, and the meeting probability of coupled random walks\cite{hamilton1982three},\cite{colding1997ricci},\cite{samal2018comparative}. In particular, it determines both the growth of ball volumes as a function of radius and the degree of overlap between two balls. A larger overlapping volume corresponds to a lower transportation cost, thereby revealing a natural connection between Ricci curvature and optimal transport theory. Building on this connection, Ollivier extended the notion of Ricci curvature to general metric spaces via 
the framework of optimal transport\cite{ollivier2007ricci, ollivier2009ricci} and further applied to graphs\cite{lin2011ricci}.  

\begin{definition}[Probability Measure]
For an undirected graph $G=(V,E)$, denote the set of neighbors of node $x\in V$ as $\mathcal{N}(x)=\{y\mid(x, y)\in E\}$, then a probability measure $m_x^\alpha$ centered at node $x$ is defined as:
\begin{equation}
m_x^\alpha(y) = 
\begin{cases}
\alpha & \text{if } y = x, \\[6pt]
\dfrac{1-\alpha}{|\mathcal{N}(x)|} & \text{if } y_i \in \mathcal{N}(x), \\[10pt]
0 & \text{otherwise},
\end{cases}
\end{equation}
where $\alpha \in [0,1]$ is a parameter that controls the weight assigned to the center node $x$. Similar to~\cite{ni2015ricci,ni2018network}, we set $\alpha=0.5$ in this study.
\end{definition}

\begin{definition}[Wasserstein Distance]
Given two probability measures $m_x$ and $m_y$ defined on the nodes $x,y\in V$ of graph $G$, the Wasserstein distance between them is defined as:
\begin{equation}
W_1(m_x, m_y) = \inf_{\pi \in \Pi(m_x, m_y)} \sum_{u,v \in V} d(u,v) \pi(u,v),
\end{equation}
where $d(u,v)$ is the shortest path distance between nodes $u$ and $v$ in graph $G$, and $\pi(u, v)$ is the amount of probability mass 
transported from node $u$ to node $v$, which is mass preserving, i.e., 
\begin{equation}
\sum_{v_j \in V} \pi(x,v_j) = m_x, \quad \sum_{u_i \in V} \pi(u_i,y) = m_y.
\end{equation} 
\end{definition}

\begin{definition}[Ollivier-Ricci Curvature]
Let $G=(V,E)$ be a graph. For any edge $(x,y) \in E$, the Ollivier-Ricci curvature $\mathrm{ORC}(x,y)$ is defined by comparing the Wasserstein distance between the probability measures $m_x$ and $m_y$ at the endpoints with their shortest path distance $d(x,y)$. Formally,
\begin{equation}
\mathrm{ORC}(x,y) = 1 - \frac{W_1(m_x, m_y)}{d(x,y)}.
\end{equation}
\end{definition}

\begin{figure}[h]
    \centering
    \includegraphics[width=1\linewidth]{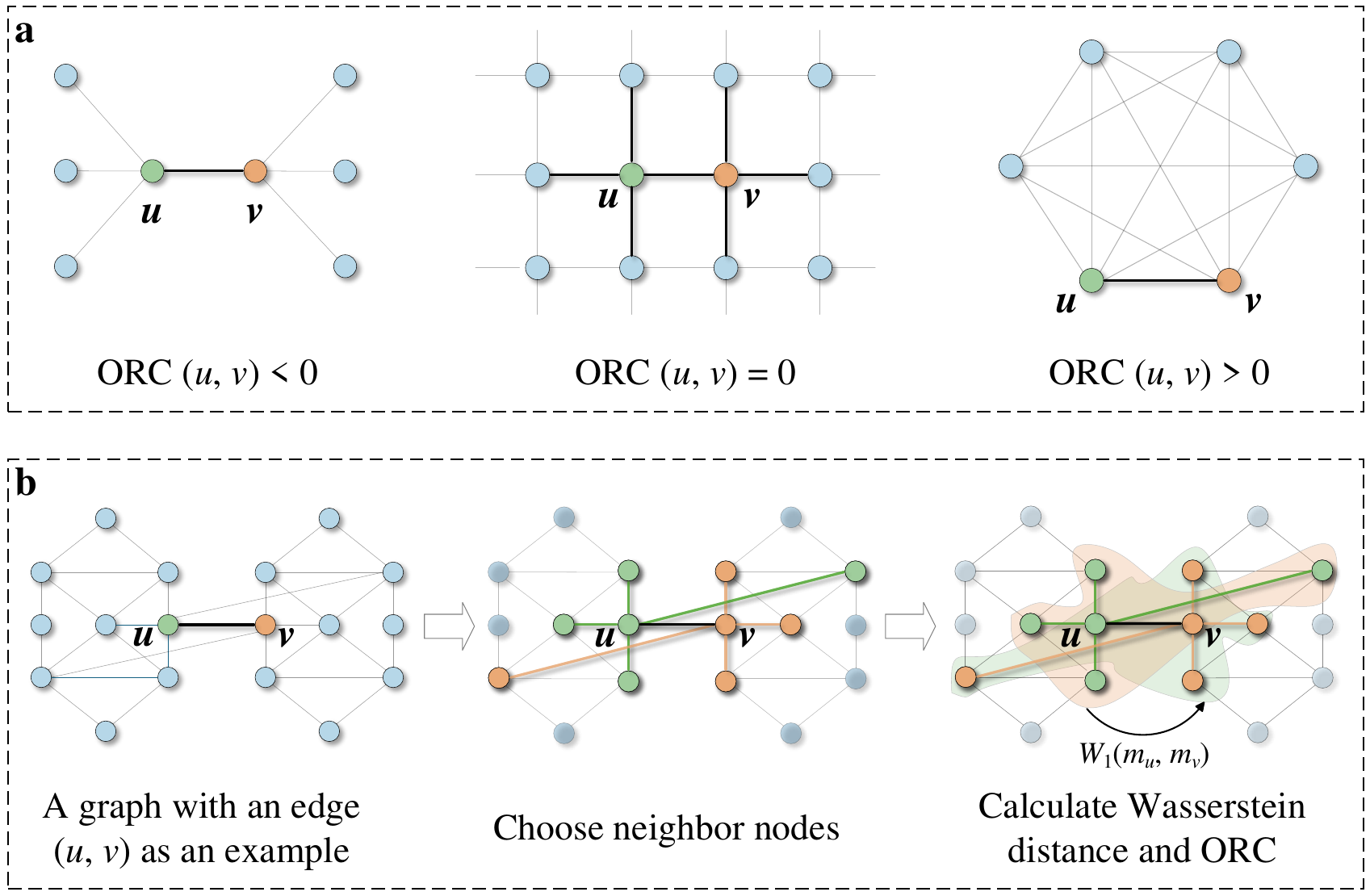}
    \caption{\textbf{An illustration of edge curvature across local graph structures and the calculation of ORC. A.} Edge curvature reflects the intrinsic geometry of local graph neighborhoods. From left to right: a tree-like structure exhibiting negative curvature, an infinitely extended grid with zero curvature, and a complete graph characterized by positive curvature. \textbf{B}. Schematic diagram for calculating the ORC of an edge $(u, v)$. $\mathrm{ORC}(u, v)$ is defined through the Wasserstein distance between two probability measures supported on the neighborhoods of nodes $u$ and $v$. The green node $u$ is connected to its green neighbors and to the orange node $v$; symmetrically, the orange node $v$ is connected to its orange neighbors and to the green node $u$.}
    \label{fig:ORC}
\end{figure}

The ORC provides a principled approach for capturing intricate topological information in graphs. Intuitively, the ORC of an 
edge reflects the relationship between the neighborhoods of its two incident nodes. As shown in Fig. \ref{fig:ORC}A, a positive ORC indicates strong connectivity, signifying that the neighboring nodes of the two vertices tend to coalesce into relatively dense local structures, thereby implying close structural relationships. In the event that the majority of edges within a local region exhibit positive curvature, the corresponding substructure is frequently classified as a community~\cite{sia2019ollivier}.  Conversely, a negative ORC is indicative of inadequate connectivity, manifesting as the tendency of neighboring nodes of the two vertices to diverge. This phenomenon results in the formation of relatively sparse local structures, signifying structural divergence.

\subsection{The RicciBind Framework}

In this section, the RicciBind framework is introduced in detail. As illustrated in Fig.~\ref{fig:framework}, RicciBind consists of three main components designed to enhance the structural modeling of protein-ligand interactions through curvature-guided geometric deep learning: (1) Curvature-Aware Graph Embedding (CAGE) encodes the protein–ligand complex as a molecular graph and refines atom-level representations by incorporating ORC as a topological descriptor (Fig.~\ref{fig:framework}B); (2) Curvature-Driven Clustering (CDC) aggregates atom-level embeddings into geometrically coherent cluster-level representations based on the local structural reflected by ORC (Fig.~\ref{fig:framework}C); (3) Optimal Transport-based Cluster Matching (OTCM) treats the protein and ligand clusters as cross-domain distributions and employs entropy-regularized optimal transport to identify and align functionally critical interaction clusters, thereby enabling accurate modeling of higher-order binding patterns (Fig.~\ref{fig:framework}D).

\subsubsection{\textbf{Curvature-Aware Graph Embedding} \rm (CAGE)}
In this section, the CAGE module is designed to encode global protein-ligand complexes into expressive atom-level representations by explicitly incorporating geometric and topological information derived from ORC. By integrating curvature as a structural descriptor within the message-passing process, CAGE captures both the local geometric configuration and the topological connectivity of molecular interactions. This curvature-aware formulation enhances the network’s ability to recognize regions of high structural complexity, thereby improving the fidelity of protein–ligand representation learning.

CAGE operates as an E(n)-invariant graph neural network (EIGN) that focuses on extracting discriminative atom-level embeddings from native binding conformations. Overall, the CAGE module can be written as
\begin{eqnarray}
    \bm{H}^L = \mathrm{CAGE}(\bm{H}^0, \bm{R}, \bm{A}, \mathrm{ORC}, L)
\end{eqnarray}
where $\bm{H}^0$ and $\bm{R}$ denote the initial features (see Supplementary Table S1 for details) and 3D coordinates of the atom nodes respectively, $\bm{A}$ is the adjacency matrix of the protein-ligand complex graph, ORC represents the Ollivier-Ricci curvature of edge in complex graph, and $L$ specifies the number of layers.

Formally, for a specific layer $l$, the message-passing procedure in CAGE is defined as
\begin{eqnarray}
    \kappa_{ij} &=& \mathrm{ORC}(v_i, v_j), \\[6pt]
    \bm{m}_{ij} &=& \phi_e(\bm{h}_i^l, \bm{h}_j^l, \|\bm{r}_i - \bm{r}_j\|^2, \kappa_{ij}), \\[6pt]
    \bm{h}_i^{l+1} &=& \phi_h(\bm{h}_i^l, \sum_{j \in \mathcal{N}(i)} \bm{m}_{ij}),
\end{eqnarray}
where $\kappa_{ij}$ is the ORC associated with edge $(i,j)$, $\bm{h}_i^l$ and $\bm{h}_j^l$ denote the features of nodes $v_i$ and $v_j$ before updating, $\|\bm{r}_i - \bm{r}_j\|^2$ represents their squared Euclidean distance computed from the 3D coordinates. The $\phi_e$ and $\phi_h$ are differentiable functions that update edge-wise and node-wise representations, respectively. Node features are initialized as $\bm{h}_i^0 = \bm{W}\bm{x}_i$, where $\bm{W}$ is a trainable linear projection. After $L$ message-passing layers, the final atom-level embeddings are obtained as 
$\bm{H} = \{\bm{h}_i^L \mid v_i \in V\}$.

\subsubsection{\textbf{Curvature-Driven Clustering} \rm (CDC)}
Protein–ligand binding often arises from cooperative interactions among higher-order structural domains and functional groups, which cannot be adequately characterized from a purely atom-level perspective\cite{lim2025cheapnet}. Moreover, over-reliance on fine-grained atomic representations not only imposes substantial computational overhead but also amplifies irrelevant noise, ultimately compromising model stability and generalization.

To overcome these limitations, we introduce the Curvature-Driven Clustering (CDC) module, which incorporates ORC into a differentiable pooling framework\cite{ying2018hierarchical} to guide the learning of cluster assignment matrices. Exploiting the geometric insight that positive curvature indicates tightly coupled local node neighborhoods whereas negative curvature reveals separation tendencies, CDC adaptively strengthens aggregation within strongly connected (positively curved) regions and suppresses ineffective coupling across weakly connected (negatively curved) ones. Through this curvature-driven clustering mechanism, CDC effectively groups topologically 
coherent atomic nodes into functionally meaningful substructures while preserving sharp boundaries in sparse regions, thereby yielding higher-order representations that are both physically grounded and structurally interpretable.

To incorporate curvature-derived geometric priors into the clustering process, the CDC module first constructs a curvature-informed adjacency matrix for both protein and ligand graphs. Specifically, for each molecular graph, the curvature-weighted edge strength is defined by transforming the ORC into a normalized connectivity measure
\begin{eqnarray}
    \Tilde{a}_{ij}= \frac{1}{1 + \text e^{-\kappa_{ij}}},
\end{eqnarray}
where $\Tilde{a}_{ij}$ represents the $(i,j)$-th entry of curvature-modulated adjacency matrix $\Tilde{\bm{A}}$. The sigmoid mapping normalizes curvature values into a stable range, assigning higher weights to geometrically cohesive regions and attenuating connections across structurally divergent areas.

To further stabilize the clustering process, the curvature-modulated adjacency matrix is augmented with self-loops and symmetrically normalized to obtain the curvature-informed adjacency matrix
\begin{eqnarray}
    \bm{A}^{\rm curv} = \hat{\bm{D}}^{-1/2}\hat{\bm{A}}\hat{\bm{D}}^{-1/2}
\end{eqnarray}
where
\begin{eqnarray*}
    \hat{\bm{A}} = \Tilde{\bm{A}}+\bm{I}, \quad 
    \hat{\bm{d}}_{ii} = \sum_j \hat{a}_{ij}.
\end{eqnarray*}

With the curvature-informed adjacency matrices $\bm{A}^{\rm curv}_p$ and $\bm{A}^{\rm curv}_l$ established, the CDC module computes the curvature-guided soft cluster assignment matrices for protein and ligand nodes. Given the atom-level node embeddings $\bm{H}_p \in \mathbb{R}^{|V_p|\times d}$ and $\bm{H}_l \in \mathbb{R}^{|V_l|\times d}$, the curvature-guided soft cluster assignment matrices 
$\bm{S}^{\rm curv}_p \in \mathbb{R}^{|V_p|\times c_p}$ and $\bm{S}^{\rm curv}_l \in \mathbb{R}^{|V_l|\times c_l}$ are computed as
\begin{eqnarray}
    \bm{S}^{\rm curv}_p = {\mathrm{Softmax}}({\rm GNN}(\bm{A}^{\rm curv}_p, \bm{H}_p, \boldsymbol{\theta}_{S_p})), \\[6pt]
    \bm{S}^{\rm curv}_l = {\mathrm{Softmax}}({\rm GNN}(\bm{A}^{\rm curv}_l, \bm{H}_l, \boldsymbol{\theta}_{S_l})),
\end{eqnarray}
where $\boldsymbol{\theta}_{S_p}$ and $\boldsymbol{\theta}_{S_l}$ are learnable parameters, and $c_p$ and $c_l$ represent the predefined numbers of clusters for the protein and ligand, respectively. 

Based on the curvature-guided cluster assignment matrices, the CDC module aggregates atom-level information into curvature-aware cluster representations. The resulting cluster-level embeddings for protein and ligand, denoted as $\bm{Z}^{\rm curv}_p \in \mathbb{R}^{c_p \times d}$ and $\bm{Z}^{\rm curv}_l \in \mathbb{R}^{c_l \times d}$, together with their corresponding curvature-aware cluster adjacency matrices 
$\tilde{\bm{A}}^{\rm curv}_p \in \mathbb{R}^{c_p \times c_p}$ and $\tilde{\bm{A}}^{\rm curv}_l \in \mathbb{R}^{c_l \times c_l}$ are obtained as
\begin{eqnarray}
&\bm{Z}^{\rm curv}_p = ({\bm{S}^{\rm curv}_p})^\top\bm{H}_p,\ \tilde{\bm{A}}^{\rm curv}_p = ({\bm{S}^{\rm curv}_p})^\top\tilde{\bm{A}}_p\bm{S}^{\rm curv}_p, \\[6pt]
&\bm{Z}^{\rm curv}_l = ({\bm{S}^{\rm curv}_l})^\top\bm{H}_l,\ \tilde{\bm{A}}^{\rm curv}_l = ({\bm{S}^{\rm curv}_l})^\top\tilde{\bm{A}}_l\bm{S}^{\rm curv}_l.
\end{eqnarray}
Here, $\bm{Z}^{\rm curv}_p$ and $\bm{Z}^{\rm curv}_l$ summarize atomic embeddings into higher-level structural units, while $\tilde{\bm{A}}^{\rm curv}_p$ and $\tilde{\bm{A}}^{\rm curv}_l$ encode curvature-enhanced inter-cluster connectivity patterns, ensuring that the geometric and topological organization of the original molecular graphs is preserved after pooling.

To further refine these hierarchical representations and capture interactions across clusters, additional GNNs are applied to the cluster-level graphs:
\begin{eqnarray}
&\bm{C}_{p}^{\rm curv} = {\rm GNN}(\tilde{\bm{A}}^{\rm curv}_p, \bm{Z}^{\rm curv}_p,\boldsymbol{\theta}_{Z_p}),\\[6pt]
&\bm{C}_{l}^{\rm curv} = {\rm GNN}(\tilde{\bm{A}}^{\rm curv}_l, \bm{Z}^{\rm curv}_l,\boldsymbol{\theta}_{Z_l}),
\end{eqnarray}
where $\boldsymbol{\theta}_{Z_p}$ and $\boldsymbol{\theta}_{Z_l}$ are learnable parameters. The resulting cluster embeddings, $\bm{C}_{p}^{\rm curv}$ and $\bm{C}_{l}^{\rm curv}$ serve as refined, semantically coherent representations of the protein and ligand at the cluster level.

\subsubsection{\textbf{Optimal Transport-Based Cluster Matching} \rm (OTCM)}
To capture the correspondence between functionally critical clusters that drive protein–ligand interactions, we introduce the OTCM module. Optimal transport (OT), as a classical mathematical framework for comparing and aligning probability distributions, establishes robust and interpretable correspondences by identifying the most cost-efficient transport plan between them\cite{monge1781memoire},\cite{villani2008optimal},\cite{ren2024towards}. 

In this study, the protein and ligand are modeled as two heterogeneous structural domains, $\mathcal{D}_p$ and $\mathcal{D}_l$, where the curvature-refined clusters produced by the CDC module define discrete probability distributions over the cluster sets of each domain. Building on this representation, cross-domain interactions between protein clusters and ligand clusters are formulated as an entropy-regularized OT problem that learns an OT plan aligning structurally and functionally related clusters across domains.

Given the refined cluster embeddings $\bm{C}_{p}^{\rm curv} \in \mathbb{R}^{c_p \times d}$ and $\bm{C}_{l}^{\rm curv} \in \mathbb{R}^{c_l \times d}$ produced by the CDC module, we first project them into a shared latent space through learnable mappings $\bm{W}_p$ and $\bm{W}_l$. This ensures that both domains are represented in a united metric space, allowing for meaningful computation of cross-domain distances. 

To quantify the correspondence between protein clusters and ligand clusters, the protein-to-ligand transport cost matrix $\bm{C}_{p \rightarrow l} \in \mathbb{R}^{c_p\times c_l}$ is then defined by the squared Euclidean distance between the projected cluster embeddings
\begin{align}
    \bm{C}_{p \rightarrow l}(i,j)=\Vert f^i_p-f^j_l\Vert_2^2,
\end{align}
where $f^i_p$ and $f^j_l$ denote the projected features of the $i$-th protein cluster and the $j$-th ligand cluster, respectively. 

Let $\bm{\gamma}_{p \rightarrow l}\in \mathbb{R}_{+}^{c_p \times c_l}$ represents the transport assignment of protein-to-ligand pairwise transport cost matrix. The cross-domain matching is then formulated as an entropy-regularized Kantorovich OT problem 
\begin{eqnarray} 
\begin{aligned}
    &\min_{\bm{\gamma}_{p \rightarrow l}} \langle \bm{\gamma}_{p \rightarrow l}, \bm{C}_{p \rightarrow l} \rangle + \lambda H(\gamma_{p \rightarrow l}),\\[6pt]
   &\text{s.t.}\ \bm{\gamma}_{p \rightarrow l} \bm{1}_{c_l} = \mu,\ \bm{\gamma}_{p \rightarrow l}^\top \bm{1}_{c_p} = \nu,
\end{aligned}
\end{eqnarray}  
where $\bm{1}_{c_l}\in \mathbb{R}^{c_l}$ and $\bm{1}_{c_p}\in \mathbb{R}^{c_p}$ are all-one vector, $\mu$ and $\nu$ denote the empirical marginal distribution of the protein and ligand clusters. The entropy term is given by $H(\bm{\gamma}_{p \rightarrow l})=\sum_{i,j}\bm{\gamma}^{ij}_{p \rightarrow l}{\rm log}\bm{\gamma}^{ij}_{p \rightarrow l}$, for penalty parameter $\lambda>0$, the OT solution $\bm{\gamma}^{*}_{p \rightarrow l}$ is unique and has the closed form
\begin{eqnarray}
    \bm{\gamma}^*_{p \rightarrow l}= {\rm diag}(\mu)\mathrm{exp}\left(-\frac{\bm{C}_{p \rightarrow l}}{\lambda}\right){\rm diag}(\nu).
\end{eqnarray}

To characterize the strength of interactions from protein clusters to ligand clusters, we perform row normalization on the optimal coupling matrix to derive a matching matrix
\begin{eqnarray}
    \bm{M}_{p \rightarrow l}={\rm diag}(\bm{\gamma}^* \bm{1}_n)^{-1}\bm{\gamma}^*\in \mathbb{R}^{c_p\times c_l},
\end{eqnarray}
which defines a many-to-many correspondence between protein clusters and ligand clusters. Intuitively, larger matching weights indicate that a protein cluster exhibits a stronger focus on a ligand cluster.

Based on the matching relationship, each protein cluster performs weighted aggregation of ligand clusters information according to the learned matching weights. The protein-to-ligand interaction representation is represented as
\begin{eqnarray}
    \bm{Z}_{p \rightarrow l} = \bm{M}_{p \rightarrow l}\bm{L},
\end{eqnarray}
where $\bm{L}=\bm{W}_{p \rightarrow l}\bm{C}_l^{\rm curv}$ denotes a learnable linear transformation of ligand cluster representations. This transformation allows the model to adaptively emphasize ligand substructures that are strongly coupled with specific protein subregions.

A symmetric ligand-to-protein representation $\bm{Z}_{l \rightarrow p}$ is obtained analogously to ensure bidirectional modeling of protein-ligand interactions. Through this bidirectional OT-based matching, the OTCM module establishes a geometry-consistent alignment between protein and ligand cluster representations, capturing key cross-domain interactions and identifying inter-cluster correspondences critical for molecular binding while attenuating the influence of irrelevant or weakly interacting clusters.

\subsubsection{\textbf{Protein-Ligand Interaction Representation and Binding Affinity Prediction}}
Finally, to derive a unified representation of the protein–ligand interaction, the curvature-informed structural embeddings produced by the CDC module are integrated with the cross-domain transport features learned by the OTCM module. This fusion jointly captures (i) the intrinsic geometric–topological organization within each molecular domain and (ii) the directional interaction patterns between protein and ligand clusters. The protein-ligand interaction representation is computed as
\begin{eqnarray}
\begin{aligned}
    \bm{Z}_{\text{pro-lig}}
&= \Phi_{l \rightarrow p}\!\left( 
    \sum_{i=1}^{c_l} \bm{Z}_{l \rightarrow p}[i] 
    + \sum_{i=1}^{c_l} \bm{C}^{\rm curv}_{l}[i] 
\right) \\
&+ \Phi_{p \rightarrow l}\!\left(
    \sum_{j=1}^{c_p} \bm{Z}_{p \rightarrow l}[j]
+ \sum_{j=1}^{c_p} \bm{C}^{\rm curv}_{p}[j]
\right) .
\end{aligned}
\end{eqnarray}
where $\Phi_{l \rightarrow p}$ and $\Phi_{p \rightarrow l}$ are multilayer perceptrons (MLPs) with learnable parameters that nonlinearly fuse intra-domain curvature-aware representations with cross-domain transport-informed features.

The protein-ligand interaction representation $\bm{Z}_{\text{pro-lig}}$ is then passed to the prediction head
\begin{eqnarray}
\hat{y} = \Phi_{\text{pro-lig}}(\bm{Z}_{\text{pro-lig}}),
\end{eqnarray}
where $\Phi_{\text{pro-lig}}$ is an MLP used to regress the binding affinity.

The overall model is trained end-to-end by minimizing the mean-squared error (MSE) between the predicted and experimentally measured binding affinities
\begin{eqnarray}
    L_{\rm MSE} = \frac{1}{N}\sum_{i=1}^N(\hat{y}_i-y_i)^2,
\end{eqnarray}
where $N$ is the number of protein-ligand complex samples, $\hat{y}_i$ denotes the predicted binding affinity of the $i$-th complex, and $y_i$ represents its corresponding experimental ground truth.

By joint integration curvature-guided structural signals with OT-based cross-domain relational correspondences, RicciBind learns representations that encode both fine-grained geometric context and high-level interaction semantics, resulting in improved fidelity and interpretability in protein–ligand binding affinity prediction.

\section{Results and Discussion}\label{Results and Discussion} 

In this study, we evaluate RicciBind on diverse PLA benchmarks tasks, including binding affinity prediction and large-scale structure-based virtual screening.

All experiments were implemented using PyTorch on an NVIDIA GeForce RTX 4080 GPU with 16 GB of memory. We employed the Adam optimizer with a learning rate of $1\textbf{e}{-4}$ to update model parameters and applied a weight decay of $1\textbf{e}{-6}$ to mitigate overfitting. The batch size was set to 128, and early stopping was adopted based on the validation RMSE to ensure stable convergence.

\subsection{Data Preparation and Experimental Setup}\label{data}
For the binding affinity prediction task, we follow the experimental protocol of EHIGN~\cite{yang2024interaction}, training and validating RicciBind on the PDBbind v2016 general set. After preprocessing with RDKit, a total of 12,904 protein-ligand complexes are retained and randomly divided into training and validation sets. To assess the 
model's generalization capability, we evaluate its performance on three independent external test sets: the PDBbind v2013 core set, PDBbind v2016 core set, and PDBbind v2019 holdout set, ensuring that there is no overlap between the training, validation, and testing samples.

Furthermore, to examine the RicciBind's out-of-distribution generalization beyond PDBbind, we evaluate it on the CSAR NRC-HiQ dataset~\cite{dunbar2013csar}. After removing redundant entries overlapping with the PDBbind training set and structures that cannot be processed by RDKit, 14 complexes are retained for evaluation.

For diverse protein evaluation task, we adopt the standardized evaluation benchmark constructed by Holoprot~\cite{somnath2021multi}. Specifically, the PDBbind v2019 refined set\cite{liu2017forging} is partitioned into training, validation, and test subsets based on a 60\% protein sequence similarity threshold, thereby strictly limiting sequence homology between the training and test targets at the data level. This splitting strategy effectively mitigates the risk of information leakage arising from protein family similarity and ensures that the target proteins in the test set are markedly distinct from those in the training set in both sequence and structural characteristics.

To evaluate the applicability of RicciBind in realistic drug discovery scenarios, we follow the cold start and warm start evaluation protocols introduced by EHIGN~\cite{yang2024interaction}, examining the predictive performance of RicciBind on previously unseen protein targets as well as on ligands with substantial structural divergence. Specifically, in the cold start setting, the original PDBbind training set is partitioned based on protein sequence similarity or ligand scaffold diversity, and new training, validation, and test sets are constructed in an 8:1:1 ratio. In the protein cold start configuration, the sequences of proteins in the validation and test sets share less than 30\% sequence identity with those in the training set, ensuring substantial structural divergence between training and testing targets. In the ligand cold start setting, all ligands in the test set possess scaffolds that are not observed during training, enabling the evaluation of the model’s generalization to novel chemical structures. By contrast, in the warm start setting, the dataset is randomly partitioned, allowing proteins and ligands to appear across multiple subsets, thereby reflecting the model’s performance under relatively familiar molecular environments. Comparative analyses under these two settings provide a systematic assessment of RicciBind’s performance on both seen and unseen molecular entities, allowing a comprehensive evaluation of its predictive robustness and generalization capability in challenging and realistic protein–ligand binding scenarios.

To further assess the practical applicability of RicciBind, we conduct large-scale structure-based virtual screening (SBVS) experiments. Unlike the regression-based binding affinity prediction task, SBVS task aims to evaluate the model's capability to distinguish active ligands from decoys—molecules that share similar physicochemical properties with actives but fail to bind to the target protein. We conduct experiments on two widely adopted benchmarks, DUD-E~\cite{mysinger2012directory} and LIT-PCBA~\cite{tran2020lit}. 

The DUD-E dataset consists of 102 target proteins, 22,886 active compounds, and a large number of decoys generated from the ZINC~\cite{irwin2005zinc} database, maintaining an approximate 1:50 active-to-decoy ratio. In this study, we focus on 57 well-characterized and functionally classifiable targets spanning four major protein families: GPCRs, kinases, nuclear proteins, and proteases. To evaluate the generalization capability of RicciBind in virtual screening, we adopt a cross-family zero-shot setting: 26 targets from the nuclear protein and protease families are used for training, while 31 targets from the GPCR and kinase families are reserved for testing. This experimental design assesses the model’s ability to perform ligand screening on entirely unseen protein families, providing a stringent test of its capacity to learn transferable protein-ligand interaction patterns. The full list of training and testing targets is provided in  Supplementary Table S4.

The LIT-PCBA dataset, in contrast, derived from PubChem-based high-throughput biological assay data and constructed using an asymmetric validation embedding (AVE) strategy, which mitigates hidden data biases and minimize structural redundancy between training and testing sets. This design yields a more realistic and challenging benchmark for 
evaluating structure-based screening models. In this study, to further examine target-specific generalization, we evaluate RicciBind on three particularly challenging targets (FEN1, KAT2A, and PKM2) from LIT-PCBA, each exhibiting an extreme active-to-decoy ratio exceeding 1:1000, thereby reflecting conditions closer to real-world drug discovery pipelines.

\begin{table*}[!ht]
\begin{center}
\begin{threeparttable}
\caption{Performance Statistics Summary for Protein–Ligand Binding Affinity Predictions on Three External Test Sets}\label{tab:table1}
\centering
\begin{tabular}{llcccccc}
\toprule
\multirow{2}{*}{Model} & 
\multirow{2}{*} &
\multicolumn{2}{c}{2013 core set ($N=107$)} & 
\multicolumn{2}{c}{2016 core set ($N=285$)} & 
\multicolumn{2}{c}{2019 holdout set ($N=4366$)} \\
\cmidrule(lr){3-8}
 & & RMSE$\downarrow$ & Pearson$\uparrow$ & RMSE$\downarrow$ & Pearson$\uparrow$ & RMSE$\downarrow$ & Pearson$\uparrow$ \\
\midrule
\multirow{6}{*}{Structure-free}
&DeepDTA\cite{ozturk2018deepdta}      & 1.639 (0.026) & 0.718 (0.014) & 1.357 (0.015) & 0.785 (0.007) & 1.485 (0.023) & 0.586 (0.012) \\
&GraphDTA-GCN\cite{nguyen2021graphdta}          & 1.749 (0.062) & 0.662 (0.032) & 1.513 (0.048) & 0.719 (0.023) & 1.763 (0.039) & 0.439 (0.021) \\
&GraphDTA-GAT\cite{nguyen2021graphdta}          & 2.043 (0.029) & 0.476 (0.022) & 1.748 (0.019) & 0.594 (0.010) & 1.663 (0.027) & 0.432 (0.016) \\
&GraphDTA-GIN\cite{nguyen2021graphdta}          & 1.691 (0.124) & 0.694 (0.059) & 1.470 (0.065) & 0.743 (0.027) & 1.676 (0.032) & 0.472 (0.021) \\
&GraphDTA-GAT-GCN\cite{nguyen2021graphdta}      & 1.645 (0.085) & 0.711 (0.036) & 1.434 (0.064) & 0.754 (0.025) & 1.705 (0.075) & 0.474 (0.028) \\
&MGraphDTA\cite{yang2022mgraphdta}    & 1.680 (0.093) & 0.696 (0.046) & 1.439 (0.047) & 0.753 (0.022) & 1.553 (0.028) & 0.538 (0.013) \\
\midrule
\multirow{11}{*}{Structure-based}
&PotentialNet\cite{feinberg2018potentialnet} & 1.607 (0.027) & 0.772 (0.007) & 1.503 (0.033) & 0.772 (0.021) & 1.514 (0.028) & 0.564 (0.014) \\
&GNN-DTI\cite{lim2019predicting}      & 1.533 (0.084) & 0.767 (0.040) & 1.384 (0.031) & 0.779 (0.008) & 1.446 (0.006) & 0.614 (0.007) \\
&IGN\cite{jiang2021interactiongraphnet}          & 1.428 (0.020) & 0.807 (0.001) & 1.269 (0.030) & 0.821 (0.013) & 1.410 (0.015) & 0.630 (0.008) \\
&SchNet\cite{schutt2018schnet}       & 1.570 (0.029) & 0.754 (0.003) & 1.390 (0.023) & 0.787 (0.016) & 1.522 (0.071) & 0.560 (0.028) \\
&EGNN\cite{satorras2021n}         & 1.498 (0.025) & 0.782 (0.015) & 1.289 (0.021) & 0.824 (0.013) & 1.399 (0.013) & 0.628 (0.009) \\
&MetalProGNet\cite{jiang2023metalprognet} & 1.494 (0.027) & 0.773 (0.009) & 1.300 (0.032) & 0.815 (0.021) & 1.448 (0.021) & 0.610 (0.009) \\
&GIGN\cite{yang2023geometric}         & 1.380 (0.009) & 0.821 (0.003) & 1.190 (0.017) & 0.840 (0.011) & 1.393 (0.007) & 0.641 (0.006) \\
&SS-GNN\cite{zhang2023ss}       & 1.330 (0.011) & 0.830 (0.007) & 1.165 (0.011) & 0.846 (0.008) & 1.450 (0.006) & 0.633 (0.004) \\
&EHIGN\cite{yang2024interaction}        & 1.297 (0.026) & 0.841 (0.006) & 1.150 (0.022) & 0.854 (0.004) & 1.368 (0.021) & \uline{0.667 (0.008)} \\
&CheapNet\cite{lim2025cheapnet}     & \uline{1.262 (0.017)} & \uline{0.857 (0.004)} & \textbf{1.107 (0.011)} & \uline{0.870 (0.002)} & \uline{1.343 (0.007)} & 0.665 (0.003) \\
\cmidrule{2-8}
% \cellcolor{gray!20}
&\textbf{RicciBind} 
& \textbf{1.259 (0.006)} 
  & \textbf{0.872 (0.003)} 
  & \uline{1.128 (0.010)} 
  & \textbf{0.871 (0.002)} 
  & \textbf{1.335 (0.003)} 
  & \textbf{0.667 (0.002)} \\
\bottomrule
\end{tabular}
\begin{tablenotes}
\footnotesize
\item Note: The evaluation uses RMSE for prediction accuracy, with the average affinity values in the test sets being 6.318 for PDBBind v2013, 6.486 for PDBBind v2016 and 6.333 for PDBBind v2019. Pearson correlation coefficient evaluates the correlation of predictions with experimental affinities. An upward arrow ($\uparrow$) denotes higher scores are better and a downward arrow ($\downarrow$) denotes the reverse. The mean and standard deviation of the results from three independent runs are reported. The best performance for each metric is highlighted in \textbf{bold} and the second-best performance is \uline{underlined}.
\end{tablenotes}
\end{threeparttable}
\end{center}
\end{table*}

\subsection{RicciBind in Cross-Dataset PLA Prediction}

In the PLA prediction task, the objective is to predict the binding strength between a given pair of protein and ligand, which is a regression problem. We first evaluated RicciBind across three non-overlapping datasets from PDBbind, comparing it against a diverse set of representative baselines, including structure-free models (DeepDTA~\cite{ozturk2018deepdta}, GraphDTA~\cite{nguyen2021graphdta} and MGraphDTA~\cite{yang2022mgraphdta}) and structure-based models (PotentialNet~\cite{feinberg2018potentialnet}, GNN-DTI~\cite{lim2019predicting}, IGN~\cite{jiang2021interactiongraphnet}, 
SchNet~\cite{schutt2018schnet}, EGNN~\cite{satorras2021n}, MetalProGNet~\cite{jiang2023metalprognet}, GIGN~\cite{yang2023geometric}, SS-GNN~\cite{zhang2023ss}, EHIGN~\cite{yang2024interaction} and CheapNet~\cite{lim2025cheapnet}). Following the experimental protocol of EHIGN, all experiments are repeated three times with different random seeds to ensure statistical reliability. Model performance is assessed using the RMSE and the Pearson's correlation coefficient, with the mean and standard deviation reported across runs. The comparative results on three independent external test sets are summarized in Table \ref{tab:table1}, where baseline results are referenced from the official reports of EHIGN\cite{yang2024interaction} and CheapNet\cite{lim2025cheapnet}.

As shown in Table \ref{tab:table1}, RicciBind achieves competitive performance in cross-dataset binding affinity prediction tasks. Overall, RicciBind substantially outperforms structure-free models and achieves performance comparable to or exceeding that of representative structure-based approaches. Specifically, on the PDBbind v2013 core set, RicciBind attains a Pearson correlation coefficient of 0.872, representing a significant improvement over existing methods. On the PDBbind v2016 core set, RicciBind similarly achieves the highest Pearson correlation of 0.871, confirming its stability and robustness across different dataset. Furthermore, on the more challenging and larger-scale PDBbind v2019 holdout set, RicciBind exhibites superior overall performance, with an RMSE of 1.335 and a Pearson correlation of 0.667, indicating strong cross-dataset generalization capabilities to previously unseen protein–ligand complexes. Notably, evaluation on the PDBbind v2019 holdout set follows a temporal split scenario, where the model is trained on historical structural data and tested on newly released protein–ligand complexes. This setup more realistically simulates the predictive requirements encountered in real-world drug discovery~\cite{yang2024interaction}.

In summary, across the three PDBbind datasets, RicciBind consistently and substantially outperforms structure-free methods, highlighting the critical importance of explicitly incorporating three-dimensional molecular structural information into binding affinity modeling for accurately characterizing protein–ligand interactions. Moreover, compared with existing structure-based models, RicciBind achieved superior or at least competitive performance across all benchmarks, demonstrating strong stability and robust cross-dataset generalization. These advantages can be attributed to the curvature-aware geometric representations, which jointly capture local geometric constraints and global topological properties of molecular structures, and to the modeling of cluster-level interactions that effectively encode higher-order structural patterns driving protein–ligand binding, thereby enabling a more fine-grained representation of protein–ligand interactions.

\begin{figure*}[!htbp]
    \centering
    \includegraphics[width=1\linewidth]{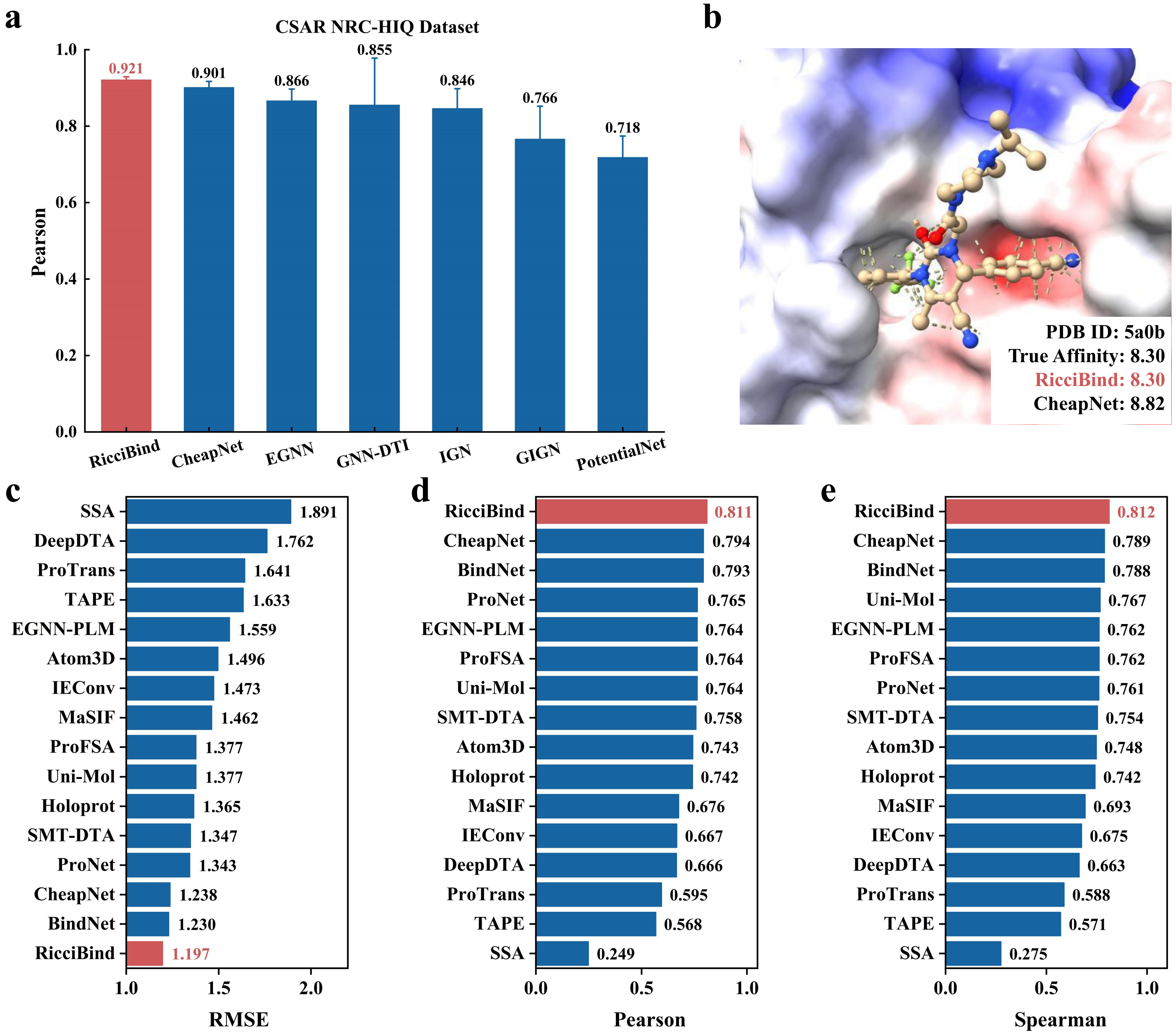}
    \caption{\textbf{Performance evaluation of RicciBind on non-PDBbind external test sets and unseen proteins.} a, Comparison of Pearson correlation coefficients obtained by training RicciBind on the PDBbind dataset and evaluating on the external CSAR NRC-HiQ test set. b, Comparative analysis of predicted binding affinities for a representative protein–ligand complex (PDB ID: 5a0b). c-e, Performance of RicciBind on unseen proteins, evaluated in terms of root mean squared error(c), Pearson correlation coefficient(d), and Spearman rank correlation coefficient. All the compared results are taken from~\cite{lim2025cheapnet}.}
    \label{fig:affinity}
\end{figure*}

\subsection{RicciBind on Non-PDBbind External Test Set}

To further assess the reliability and generalization capability of RicciBind beyond the PDBbind domain, we evaluate models trained on PDBbind dataset using the CSAR NRC-HiQ dataset\cite{dunbar2013csar}. Compared with PDBbind, CSAR NRC-HiQ adopts more stringent curation and quality-control criteria, and is therefore widely regarded as a rigorous benchmark for evaluating the practical applicability of protein–ligand binding affinity prediction models. As summarized in Fig.~\ref{fig:affinity}a, RicciBind achieves the highest Pearson correlation coefficient of 0.92 among structure-based baselines, demonstrating highly competitive performance under this challenging cross-dataset evaluation setting. These results demonstrate that RicciBind is capable of learning transferable protein–ligand interaction patterns, achieving stable predictive performance on previously unseen complex structures.

\subsection{RicciBind in Diverse Protein Dataset}

To evaluate the generalization capability of RicciBind in characterising the structures of unknown target proteins under substantially increased target protein structural diversity, we adopted the standardized evaluation benchmark constructed by Holoprot (see Subsection~\ref{data} for details of the dataset).

The comparison results are summarized in Fig.~\ref{fig:affinity}c-e, with baseline performances referenced from CheapNet~\cite{lim2025cheapnet}, ProNet~\cite{wang2022learning} and BindNet~\cite{feng2023protein}. RicciBind consistently outperforms all baselines across RMSE (Fig.~\ref{fig:affinity}c), Pearson's correlation coefficient (Fig.~\ref{fig:affinity}d), and Spearman rank correlation coefficient (Fig.~\ref{fig:affinity}e). These results demonstrate its clear advantages in both predictive accuracy and ranking consistency. Specifically, RicciBind achieved an RMSE of 1.197, a Pearson correlation of 0.811, and a Spearman correlation of 0.812, representing relative improvements of 3.3\%, 2.1\%, and 2.9\% over the strongest baseline, respectively.

Notably, as a structure-based hierarchical higher-order geometric representation method, RicciBind significantly outperforms models that rely primarily on protein surface information (e.g., MaSIF\cite{gainza2020deciphering}) as well as structure-based molecular modeling approaches (e.g., BindNet\cite{feng2023protein}). These results suggest that, when target protein structural diversity is substantially increased, ORC provides an effective geometric prior for jointly encoding fine-grained molecular structure and higher-order interaction patterns. This design contributes to the strong generalization ability and predictive stability of RicciBind across diverse and unseen target proteins (see Supplementary Table S2 for details).

\begin{table*}[!t]
\caption{Performance Evaluation for Structure-Base Models in the Warm Start (Random Split) and Cold Start (Split Based on Sequence Identity and Scaffold) Scenarios}\label{tab:table2}
\centering
\begin{tabular}{lcccccc}
\toprule
\multirow{2}{*}{Model} & 
\multicolumn{2}{c}{Random} & 
\multicolumn{2}{c}{Sequence identity(30\%)} & 
\multicolumn{2}{c}{Scaffold} \\
% \cmidrule(lr){2-3} \cmidrule(lr){4-5} \cmidrule(lr){6-7}
\cmidrule(lr){2-7}
& RMSE$\downarrow$ & Pearson$\uparrow$ & RMSE$\downarrow$ & Pearson$\uparrow$ & RMSE$\downarrow$ & Pearson$\uparrow$ \\
\midrule
RF-Score~\cite{ballester2010machine}     & 1.374 (0.002) & 0.680 (0.001) & 1.469 (0.001)       & 0.582 (0.001)       & 1.385 (0.002)   & 0.637 (0.001)   \\
PotentialNet~\cite{feinberg2018potentialnet}  & 1.360 (0.030) & 0.691 (0.013) & 1.582 (0.017)       & 0.507 (0.006)       & 1.538 (0.006)   & 0.532 (0.015) \\
GNN-DTI~\cite{lim2019predicting}       & 1.403 (0.011) & 0.666 (0.007) & 1.499 (0.008)       & 0.562 (0.006)       & 1.401 (0.027)   & 0.631 (0.018) \\
IGN~\cite{jiang2021interactiongraphnet}          & 1.295 (0.011) & 0.732 (0.006) & 1.509 (0.034)       & 0.561 (0.025)       & 1.361 (0.041)   & 0.665 (0.023) \\
SchNet~\cite{schutt2018schnet}       & 1.398 (0.044) & 0.672 (0.021) & 1.499 (0.022)       & 0.566 (0.020)       & 1.456 (0.046)   & 0.607 (0.014) \\
EGNN~\cite{satorras2021n}         & 1.285 (0.025) & 0.730 (0.012) & 1.483 (0.029)       & 0.579 (0.017)       & 1.395 (0.036)   & 0.635 (0.029) \\
MetalProGNet~\cite{jiang2023metalprognet} & 1.307 (0.011) & 0.729 (0.003) & 1.500 (0.029)       & 0.574 (0.013)       & 1.397 (0.028)   & 0.654 (0.009) \\
GIGN ~\cite{yang2023geometric}        & 1.264 (0.015) & 0.746 (0.006) & 1.521 (0.012)       & 0.565 (0.007)       & 1.353 (0.002)   & 0.685 (0.004) \\
SS-GNN ~\cite{zhang2023ss}      & 1.353 (0.005) & 0.714 (0.004) & 1.539 (0.031)       & 0.573 (0.023)       & 1.379 (0.037)   & 0.678 (0.017) \\
EHIGN ~\cite{yang2024interaction}      & \uline{1.248 (0.022)} & \uline{0.753 (0.007)} & \uline{1.449 (0.011)}       & \uline{0.613 (0.009)}       & \uline{1.292 (0.016)}   & \uline{0.710 (0.008)} \\
\midrule
\textbf{RicciBind} & \textbf{1.214 (0.012)} & \textbf{0.763 (0.006)} & \textbf{1.437 (0.008)} & \textbf{0.622 (0.002)} & \textbf{1.248 (0.009)} & \textbf{0.722 (0.001)} \\
\bottomrule
\end{tabular}
\end{table*} 

\subsection{RicciBind in Cold Start Scenario}

In real-world drug discovery, PLA prediction models are expected to exhibit strong generalization capabilities, enabling reliable predictions for proteins or ligands that were not observed during training and that exhibit substantial structural differences from known samples. To evaluate RicciBind under such challenging generalization scenarios, we follow the experimental protocol proposed in EHIGN\cite{yang2024interaction} and assessed its performance under both cold start and warm start settings (see Subsection~\ref{data} for details of the dataset).

Table \ref{tab:table2} summarizes the performance of RicciBind under both warm start and cold start evaluation scenarios. Overall, RicciBind achieves stable and consistent advantages across all three data partitioning strategies, outperforming the SOTA structure-based method EHIGN\cite{yang2024interaction}. This highlights its reliability and robustness in challenging prediction settings, with particularly notable generalization performance on unseen protein–ligand complexes. Specifically, in the warm start setting, RicciBind achieves the lowest RMSE and the highest Pearson correlation coefficient among all competing methods, yielding relative improvements of 2.7\% and 1.3\%, respectively. This indicates that when the test distribution is closely aligned with the training data, RicciBind can more effectively exploit structural information to finely characterize protein–ligand binding patterns, resulting in improved predictive accuracy and stability. By contrast, the cold start setting poses more stringent challenges to model generalization. In both the protein cold start split based on a 30\% sequence identity threshold and the ligand cold start split based on scaffold dissimilarity, RicciBind maintained its leading performance, achieving the lowest RMSE values (1.437 and 1.248) and the highest Pearson correlation coefficients (0.622 and 0.722), respectively. This demonstrates that RicciBind remains capable of effectively learning and capturing key protein–ligand interaction patterns even under conditions of pronounced structural variation in either proteins or ligands.  

Notably, under cold start scenarios, all models exhibit varying degrees of performance degradation, highlighting the persistent challenge of improving generalization to entirely unseen structures. Nevertheless, RicciBind consistently achieves the best average performance and the lowest standard deviation across all evaluation scenarios, demonstrating its robustness and predictive stability under diverse data distributions. Collectively, these results suggest that RicciBind not only effectively learns geometric characteristics from previously observed molecules, but also retains the ability to extract key interaction patterns when substantial structural variations occur. This reflects the potential advantages of its curvature-guided geometric characterization and cluster-level interaction modeling mechanism in mitigating distribution shifts caused by structural differences, thereby supporting more robust and reliable PLA prediction in challenging generalization settings.   

\subsection{RicciBind in Structure-Based Virtual Screening}

To assess the practical applicability of RicciBind in real-world drug discovery pipelines, we conduct comprehensive experiments on two structure-based virtual screening (SBVS) benchmarks. Virtual screening aims to prioritize the identification of potentially active molecules from large-scale compound libraries by leveraging protein-ligand 3D structural information. Its core challenge lies in the model's ability to identify true binders at an early stage. Following the standard evaluation protocols, we employed the enrichment factor (EF) as the primary performance metric and reported EF values at the top 0.1\%, 0.5\%, 1\%, and 5\% of ranked compound lists, enabling a quantitative comparison of the early enrichment capability across different methods.

\begin{figure}[!htbp]
    \centering
    \includegraphics[width=1\linewidth]{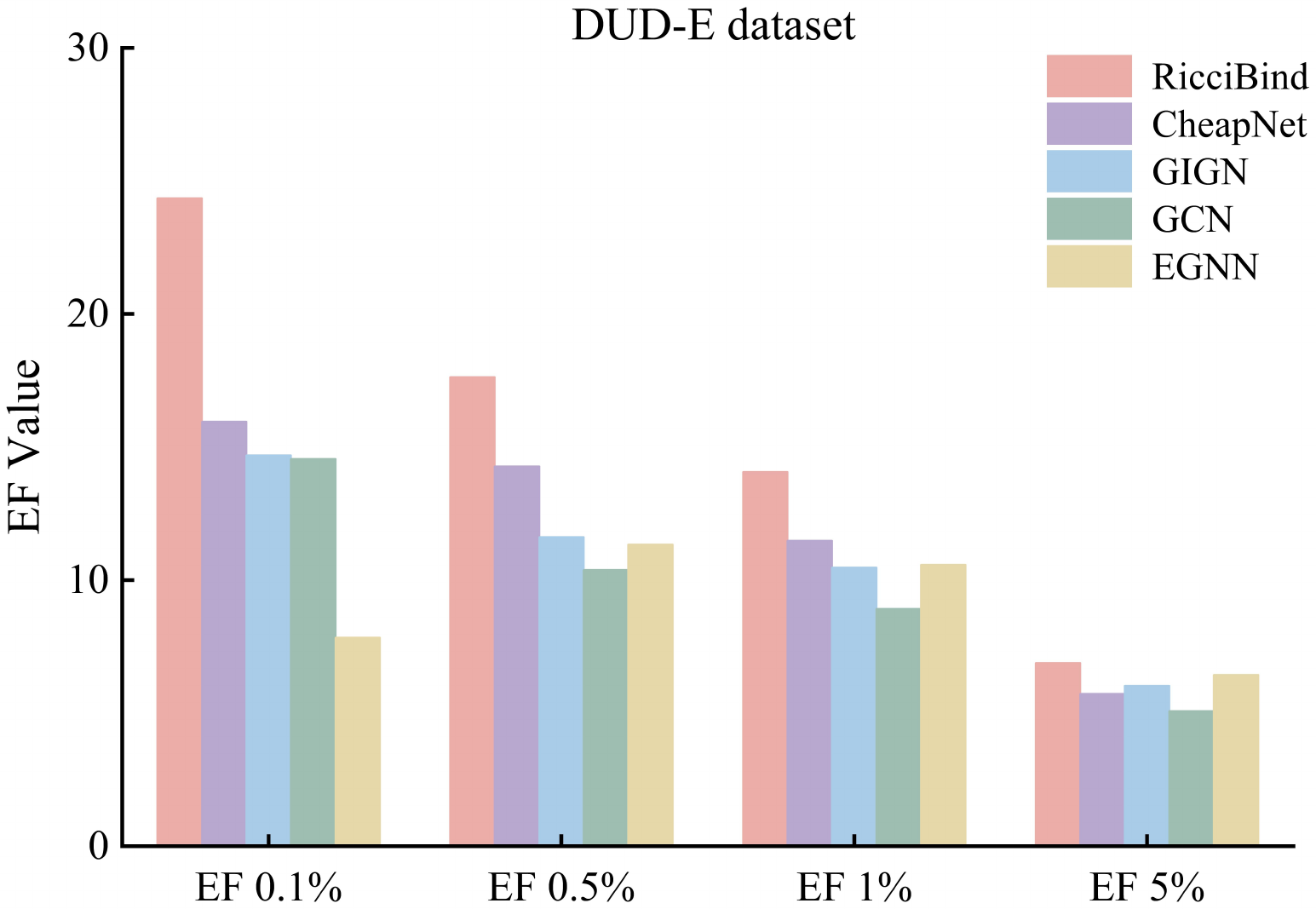}
    \caption{Performance Comparison of Different Models on the DUD-E Dataset.}
    \label{fig:DUDE}
\end{figure}

We first evaluated the virtual screening performance of RicciBind on a high-confidence subset of the DUD-E benchmark. The assessment was conducted under a stringent cross-family zero-shot setting, designed to test the RicciBind’s generalization capability towards unknown targets under conditions approximating real-world drug discovery scenarios. As shown in  Fig. \ref{fig:DUDE}, RicciBind consistently outperforms 
all competing methods across four evaluation thresholds (top 0.1\%, 0.5\%, 1\%, and 5\%). Notably, under the most stringent early enrichment criteria, RicciBind achieves enrichment factors of 24.346 and 17.630 at the top 0.1\% and 0.5\% thresholds, respectively, indicating its strong ability to screen out genuinely active compounds from a vast array of candidate molecules. These results demonstrate the robust generalization of RicciBind under cross-family zero-shot conditions and highlight its potential value for early hit identification in practical high-throughput virtual screening tasks (see Subsection~\ref{data} and Supplementary Table S3 for details).

Subsequently, We conduct target-specific SBVS experiments on the LIT-PCBA benchmark, focusing on three well-recognized and particularly challenging targets, namely FEN1, KAT2A, and PKM2, to further validate the robustness and generalization capability of RicciBind  in complex screening scenarios. In this setting, RicciBind was compared against docking-based scoring functions and learning-based approaches. Relevant benchmark results were uniformly sourced from EHIGN\cite{yang2024interaction} to ensure consistency in evaluation. As illustrated in Table \ref{tab:table4}, traditional docking scoring functions (e.g., Glide SP\cite{friesner2004glide}) exhibit inferior screening performance across all three targets compared with learning-based methods. This highlights the challenges posed by complex structure-activity relationships in the LIT-PCBA dataset to purely physical scoring models, while also revealing the inherent limitations of docking-based approaches in early enrichment tasks. In contrast, RicciBind achieves competitive and stable enrichment performance across all three targets. Specifically, for the KAT2A target, RicciBind achieves a improvement of approximately 50\% over the current SOTA structure-based model EHIGN on the EF${0.1\%}$ metric, indicating its enhanced ability to effectively identify active compounds across diverse and challenging screening scenarios.

Overall, the evaluations on cross-family zero-shot virtual screening using DUD-E and target-specific virtual screening on LIT-PCBA demonstrate that RicciBind maintains robust and superior early enrichment performance even under screening scenarios characterised by high chemical space diversity, structurally complex protein targets, and pronounced distributional shifts. These results underscore the practical utility and translational potential of RicciBind for real-world structure-based drug discovery applications.

\begin{table}[!htbp]
\centering
\caption{Target-Specific SBVS on the LIT-PCBA Dataset}\label{tab:table4}
\resizebox{0.48\textwidth}{!}{
\begin{tabular}{llcccc}
\midrule
\textbf{Target} & \textbf{Model} & \textbf{EF$0.1\%$ $\uparrow$} & \textbf{EF$0.5\%$ $\uparrow$} & \textbf{EF$1\%$ $\uparrow$} & \textbf{EF$5\%$ $\uparrow$} \\
\midrule
\multirow{8}{*}{FEN1} 
& IFP~\cite{marcou2007optimizing} & - & - & 6.78 & - \\
& GRIM~\cite{desaphy2013encoding} & - & - & 7.32 & - \\
& Surflex~\cite{jain2007surflex} & - & - & 4.34 & - \\
& Glide SP~\cite{friesner2004glide} & 0.00 & 2.17 & 2.17 & 4.13 \\
& RFscore-VS~\cite{wojcikowski2017performance} & 65.05 & 26.08 & 14.13 & 5.22 \\
& IGN\cite{jiang2021interactiongraphnet} & 119.26 & 41.29 & 30.42 & 10.65 \\
& EHIGN\cite{yang2024interaction} & \uline{154.57} & \uline{75.41} & \textbf{52.18} & \textbf{14.22} \\
\cmidrule{2-6}
& \textbf{RicciBind} & \textbf{165.61} & \textbf{77.63} & \uline{42.19} & \uline{13.11} \\
\midrule
\multirow{8}{*}{KAT2A} 
& IFP~\cite{marcou2007optimizing} & - & - & 3.61 & - \\
& GRIM~\cite{desaphy2013encoding} & - & - & 3.61 & - \\
& Surflex~\cite{jain2007surflex} & - & - & 2.58 & - \\
& Glide SP~\cite{friesner2004glide} & 0.00 & 0.00 & 0.00 & 0.83 \\
& RFscore-VS~\cite{wojcikowski2017performance} & 0.00 & 0.00 & 2.08 & 1.25 \\
& IGN\cite{jiang2021interactiongraphnet} & 0.00 & 4.16 & 4.16 & 1.67 \\
& EHIGN\cite{yang2024interaction} & \uline{41.47} & \uline{16.67} & \textbf{16.67} & \uline{4.58} \\
\cmidrule{2-6}
& \textbf{RicciBind} & \textbf{62.21} & \textbf{20.83} & \uline{12.50} & \textbf{5.00} \\
\midrule
\multirow{8}{*}{PKM2} 
& IFP~\cite{marcou2007optimizing} & - & - & 7.14 & - \\
& GRIM~\cite{desaphy2013encoding} & - & - & 5.86 & - \\
& Surflex~\cite{jain2007surflex} & - & - & 0.18 & - \\
& Glide SP~\cite{friesner2004glide} & 0.00 & 0.00 & 0.00 & 0.97 \\
& RFscore-VS~\cite{wojcikowski2017performance} & 8.30 & 1.66 & 1.66 & 1.83 \\
& IGN\cite{jiang2021interactiongraphnet} & 0.00 & 3.32 & 2.50 & 3.50 \\
& EHIGN\cite{yang2024interaction} & \textbf{21.78} & \uline{7.33} & \uline{8.08} & \textbf{4.56} \\
\cmidrule{2-6}
&\textbf{RicciBind} &\textbf{21.78} &\textbf{11.73} &\textbf{9.55} &\uline{4.26} \\
\midrule
\end{tabular}}
\end{table}

\subsection{Ablation Study}

To validate the critical role of ORC in capturing the geometric information of protein-ligand interactions, we conduct a series of ablation studies to evaluate the impact of removing ORC information on RicciBind performance within both the CAGE and CDC modules. Specifically, we designed three RicciBind variants to analyze the contribution of Ricci curvature features across different modules, thereby elucidating the significance of curvature-guided geometric representations in protein-ligand interaction modeling.

\begin{table*}[ht]
\caption{Results of RicciBind with varying settings on the 2013 core set, 2016 core set, and the 2019 holdout set.}
\label{tab:table5}
\centering
\resizebox{\textwidth}{!}{
\begin{tabular}{lccccccccc}
\toprule
\multirow{2}{*}{Model} & 
\multicolumn{3}{c}{2013 Core Set} & 
\multicolumn{3}{c}{2016 Core Set} & 
\multicolumn{3}{c}{2019 Holdout Set} \\
\cmidrule(lr){2-4} \cmidrule(lr){5-7} \cmidrule(lr){8-10}
& RMSE$\downarrow$ & Pearson$\uparrow$ & Spearman$\uparrow$ 
& RMSE$\downarrow$ & Pearson$\uparrow$ & Spearman$\uparrow$ 
& RMSE$\downarrow$ & Pearson$\uparrow$ & Spearman$\uparrow$ \\
\midrule
Variant 1 & 1.351 (0.029) & 0.834 (0.010) & 0.839 (0.012)
          & 1.213 (0.048) & 0.841 (0.017) & 0.838 (0.015)
          & 1.361 (0.026) & 0.652 (0.015) & 0.639 (0.016) \\
Variant 2 & 1.319 (0.027) & 0.846 (0.008) & 0.851 (0.000) 
          & 1.143 (0.019) & 0.863 (0.006) & 0.864 (0.006)
          & 1.343 (0.016) & 0.663 (0.008) & 0.652 (0.008) \\
Variant 3 & 1.330 (0.045) & 0.850 (0.016) & 0.849 (0.021) 
          & 1.192 (0.026) & 0.851 (0.011) & 0.845 (0.015)
          & 1.354 (0.008) & 0.654 (0.004) & 0.641 (0.003) \\
\textbf{RicciBind} & \textbf{1.259 (0.006)} & \textbf{0.872 (0.003)} & \textbf{0.874 (0.007)} 
          & \textbf{1.128 (0.010)} & \textbf{0.871 (0.002)} & \textbf{0.870 (0.001)}
          & \textbf{1.335 (0.003)} & \textbf{0.667 (0.002)} & \textbf{0.653 (0.003)} \\
\bottomrule
\end{tabular}
}
\end{table*} 

\begin{itemize}
\item \textbf{Variant 1}: ORC information is removed from both the CAGE and CDC modules. This variant evaluates the overall contribution of curvature to protein–ligand interaction modeling in RicciBind.
\item \textbf{Variant 2}: ORC information is removed only from the CAGE module, while the complex graph node embeddings are encoded using the standard EIGN framework. This configuration investigates the role of curvature in the explicitly modeling higher-order topological and geometric features within the interaction graph.
\item \textbf{Variant 3}: ORC information is removed only from the CDC module, and conventional DiffPool is used to cluster protein and ligand nodes. This variant assesses the importance of curvature-guided clustering for capturing local compactness and geometrically cohesive substructures in molecular graphs.
\end{itemize}   

We evaluated the above RicciBind variants on cross-dataset PLA prediction task, with the results summarized in Table \ref{tab:table5}. Overall, all three ablation variants exhibited varying degrees of performance degradation compared with the full RicciBind model during cross-dataset testing, underscoring the importance of curvature information in characterizing protein–ligand interactions. In particular, Variant 1, which completely removes curvature information from both the CAGE and CDC modules, suffers the most pronounced performance decline across all evaluation metrics. This indicates that without curvature-guided geometric representations, the model struggles to adequately capture the intricate structural patterns inherent in protein–ligand interactions, highlighting the central role of curvature information within RicciBind. By contrast, Variant 2 and Variant 3 retain curvature information in the CDC and CAGE modules, respectively, and both achieve marked performance recovery relative to Variant 1. This suggests that curvature-guided geometric representations not only effectively enhances the model's ability to capture higher-order topological relationships but also significantly improves its geometric sensitivity, predictive stability, and cross-dataset generalization performance.

\section{Conclusion}\label{Conclusion}
The curvature-guided geometric deep learning framework, called RicciBind, is proposed to address the challenge of effectively modeling the complex geometric and topological structures underlying molecular interactions. It integrates Ollivier Ricci curvature–driven hierarchical geometric representations with an optimal transport–based structural matching mechanism to model protein–ligand interactions and enhance binding affinity prediction. Extensive experiments across multiple benchmark datasets demonstrate that RicciBind consistently exhibits competitive predictive performance and strong generalization capabilities in binding affinity prediction and large-scale structure-based virtual screening tasks. This indicates that integrating curvature-aware geometric representations with optimal transport theory provides an effective strategy for modeling molecular interactions and predicting protein-ligand binding affinities with high accuracy, thereby offering a promising new paradigm for structure-based drug discovery.

RicciBind focuses on capturing protein–ligand interaction representations, its underlying modeling paradigm is highly extensible and has the potential to generalize to broader biomolecular representations and related tasks. In the future, we plan to refine RicciBind’s modeling strategies and explore its applicability to a wider range of tasks, such as pocket detection and molecular docking, thereby advancing the drug development process.

\section*{References}
\bibliographystyle{IEEEtran}
\bibliography{references}

\begin{thebibliography}{10}
\providecommand{\url}[1]{#1}
\csname url@samestyle\endcsname
\providecommand{\newblock}{\relax}
\providecommand{\bibinfo}[2]{#2}
\providecommand{\BIBentrySTDinterwordspacing}{\spaceskip=0pt\relax}
\providecommand{\BIBentryALTinterwordstretchfactor}{4}
\providecommand{\BIBentryALTinterwordspacing}{\spaceskip=\fontdimen2\font plus
\BIBentryALTinterwordstretchfactor\fontdimen3\font minus \fontdimen4\font\relax}
\providecommand{\BIBforeignlanguage}[2]{{%
\expandafter\ifx\csname l@#1\endcsname\relax
\typeout{** WARNING: IEEEtran.bst: No hyphenation pattern has been}%
\typeout{** loaded for the language `#1'. Using the pattern for}%
\typeout{** the default language instead.}%
\else
\language=\csname l@#1\endcsname
\fi
#2}}
\providecommand{\BIBdecl}{\relax}
\BIBdecl

\bibitem{fleming2018artificial}
N.~Fleming, ``How artificial intelligence is changing drug discovery,'' \emph{Nature}, vol. 557, no. 7706, pp. S55--S57, 2018.

\bibitem{stark2022equibind}
H.~St{\"a}rk, O.~Ganea, L.~Pattanaik, R.~Barzilay, and T.~Jaakkola, ``Equibind: Geometric deep learning for drug binding structure prediction,'' in \emph{International Conference on Machine Learning}.\hskip 1em plus 0.5em minus 0.4em\relax PMLR, 2022, pp. 20\,503--20\,521.

\bibitem{chen2024multiscale}
D.~Chen, J.~Liu, and G.-W. Wei, ``Multiscale topology-enabled structure-to-sequence transformer for protein--ligand interaction predictions,'' \emph{Nature Machine Intelligence}, vol.~6, no.~7, pp. 799--810, 2024.

\bibitem{ozturk2018deepdta}
H.~{\"O}zt{\"u}rk, A.~{\"O}zg{\"u}r, and E.~Ozkirimli, ``Deepdta: deep drug--target binding affinity prediction,'' \emph{Bioinformatics}, vol.~34, no.~17, pp. i821--i829, 2018.

\bibitem{nguyen2021graphdta}
T.~Nguyen, H.~Le, T.~P. Quinn, T.~Nguyen, T.~D. Le, and S.~Venkatesh, ``Graphdta: predicting drug--target binding affinity with graph neural networks,'' \emph{Bioinformatics}, vol.~37, no.~8, pp. 1140--1147, 2021.

\bibitem{yang2023geometric}
Z.~Yang, W.~Zhong, Q.~Lv, T.~Dong, and C.~Yu-Chian~Chen, ``Geometric interaction graph neural network for predicting protein--ligand binding affinities from 3d structures (gign),'' \emph{The Journal of Physical Chemistry Letters}, vol.~14, no.~8, pp. 2020--2033, 2023.

\bibitem{shen2024curvature}
C.~Shen, P.~Ding, J.~Wee, J.~Bi, J.~Luo, and K.~Xia, ``Curvature-enhanced graph convolutional network for biomolecular interaction prediction,'' \emph{Computational and Structural Biotechnology Journal}, vol.~23, pp. 1016--1025, 2024.

\bibitem{torng2019graph}
W.~Torng and R.~B. Altman, ``Graph convolutional neural networks for predicting drug-target interactions,'' \emph{Journal of Chemical Information and Modeling}, vol.~59, no.~10, pp. 4131--4149, 2019.

\bibitem{li2021co}
T.~Li, X.-M. Zhao, and L.~Li, ``Co-vae: Drug-target binding affinity prediction by co-regularized variational autoencoders,'' \emph{IEEE Transactions on Pattern Analysis and Machine Intelligence}, vol.~44, no.~12, pp. 8861--8873, 2021.

\bibitem{bai2023interpretable}
P.~Bai, F.~Miljkovi{\'c}, B.~John, and H.~Lu, ``Interpretable bilinear attention network with domain adaptation improves drug--target prediction,'' \emph{Nature Machine Intelligence}, vol.~5, no.~2, pp. 126--136, 2023.

\bibitem{gomes2017atomic}
J.~Gomes, B.~Ramsundar, E.~N. Feinberg, and V.~S. Pande, ``Atomic convolutional networks for predicting protein-ligand binding affinity,'' \emph{arXiv preprint arXiv:1703.10603}, 2017.

\bibitem{jimenez2018k}
J.~Jim{\'e}nez, M.~Skalic, G.~Martinez-Rosell, and G.~De~Fabritiis, ``K deep: protein--ligand absolute binding affinity prediction via 3d-convolutional neural networks,'' \emph{Journal of Chemical Information and Modeling}, vol.~58, no.~2, pp. 287--296, 2018.

\bibitem{jiang2021interactiongraphnet}
D.~Jiang, C.-Y. Hsieh, Z.~Wu, Y.~Kang, J.~Wang, E.~Wang, B.~Liao, C.~Shen, L.~Xu, J.~Wu \emph{et~al.}, ``Interactiongraphnet: a novel and efficient deep graph representation learning framework for accurate protein--ligand interaction predictions,'' \emph{Journal of Medicinal Chemistry}, vol.~64, no.~24, pp. 18\,209--18\,232, 2021.

\bibitem{feinberg2018potentialnet}
E.~N. Feinberg, D.~Sur, Z.~Wu, B.~E. Husic, H.~Mai, Y.~Li, S.~Sun, J.~Yang, B.~Ramsundar, and V.~S. Pande, ``Potentialnet for molecular property prediction,'' \emph{ACS Central Science}, vol.~4, no.~11, pp. 1520--1530, 2018.

\bibitem{yang2024interaction}
Z.~Yang, W.~Zhong, Q.~Lv, T.~Dong, G.~Chen, and C.~Y.-C. Chen, ``Interaction-based inductive bias in graph neural networks: enhancing protein-ligand binding affinity predictions from 3d structures,'' \emph{IEEE Transactions on Pattern Analysis and Machine Intelligence}, vol.~46, no.~12, pp. 8191--8208, 2024.

\bibitem{zhang2023planet}
X.~Zhang, H.~Gao, H.~Wang, Z.~Chen, Z.~Zhang, X.~Chen, Y.~Li, Y.~Qi, and R.~Wang, ``Planet: a multi-objective graph neural network model for protein--ligand binding affinity prediction,'' \emph{Journal of Chemical Information and Modeling}, vol.~64, no.~7, pp. 2205--2220, 2023.

\bibitem{chen2025local}
S.~Chen, H.~Yi, Z.~You, X.~Shang, Y.-A. Huang, L.~Wang, and Z.~Wang, ``Local--global structure-aware geometric equivariant graph representation learning for predicting protein--ligand binding affinity,'' \emph{IEEE Transactions on Neural Networks and Learning Systems}, 2025.

\bibitem{du2023new}
Y.~Du, L.~Wang, D.~Feng, G.~Wang, S.~Ji, C.~P. Gomes, Z.-M. Ma \emph{et~al.}, ``A new perspective on building efficient and expressive 3d equivariant graph neural networks,'' \emph{Advances in Neural Information Processing Systems}, vol.~36, pp. 66\,647--66\,674, 2023.

\bibitem{kong2024generalist}
X.~Kong, W.~Huang, and Y.~Liu, ``Generalist equivariant transformer towards 3d molecular interaction learning,'' in \emph{International Conference on Machine Learning}.\hskip 1em plus 0.5em minus 0.4em\relax PMLR, 2024, pp. 25\,149--25\,175.

\bibitem{lim2025cheapnet}
H.~Lim, S.~Kim, and S.~Lee, ``Cheapnet: Cross-attention on hierarchical representations for efficient protein-ligand binding affinity prediction,'' in \emph{The Thirteenth International Conference on Learning Representations}, 2025.

\bibitem{ying2018hierarchical}
Z.~Ying, J.~You, C.~Morris, X.~Ren, W.~Hamilton, and J.~Leskovec, ``Hierarchical graph representation learning with differentiable pooling,'' \emph{Advances in Neural Information Processing Systems}, vol.~31, 2018.

\bibitem{ollivier2009ricci}
Y.~Ollivier, ``Ricci curvature of markov chains on metric spaces,'' \emph{Journal of Functional Analysis}, vol. 256, no.~3, pp. 810--864, 2009.

\bibitem{forman2003bochner}
Forman, ``Bochner's method for cell complexes and combinatorial ricci curvature,'' \emph{Discrete \& Computational Geometry}, vol.~29, no.~3, pp. 323--374, 2003.

\bibitem{ni2019community}
C.-C. Ni, Y.-Y. Lin, F.~Luo, and J.~Gao, ``Community detection on networks with ricci flow,'' \emph{Scientific Reports}, vol.~9, no.~1, p. 9984, 2019.

\bibitem{sia2019ollivier}
J.~Sia, E.~Jonckheere, and P.~Bogdan, ``Ollivier-ricci curvature-based method to community detection in complex networks,'' \emph{Scientific Reports}, vol.~9, no.~1, p. 9800, 2019.

\bibitem{ni2015ricci}
C.-C. Ni, Y.-Y. Lin, J.~Gao, X.~D. Gu, and E.~Saucan, ``Ricci curvature of the internet topology,'' in \emph{IEEE Conference on Computer Communications (INFOCOM)}.\hskip 1em plus 0.5em minus 0.4em\relax IEEE, 2015, pp. 2758--2766.

\bibitem{wee2021forman}
J.~Wee and K.~Xia, ``Forman persistent ricci curvature (fprc)-based machine learning models for protein--ligand binding affinity prediction,'' \emph{Briefings in Bioinformatics}, vol.~22, no.~6, 2021.

\bibitem{wee2021ollivier}
------, ``Ollivier persistent ricci curvature-based machine learning for the protein--ligand binding affinity prediction,'' \emph{Journal of Chemical Information and Modeling}, vol.~61, no.~4, pp. 1617--1626, 2021.

\bibitem{nguyen2023revisiting}
K.~Nguyen, N.~M. Hieu, V.~D. Nguyen, N.~Ho, S.~Osher, and T.~M. Nguyen, ``Revisiting over-smoothing and over-squashing using ollivier-ricci curvature,'' in \emph{International Conference on Machine Learning}.\hskip 1em plus 0.5em minus 0.4em\relax PMLR, 2023, pp. 25\,956--25\,979.

\bibitem{ye2019curvature}
Z.~Ye, K.~S. Liu, T.~Ma, J.~Gao, and C.~Chen, ``Curvature graph network,'' in \emph{International Conference on Learning Representations}, 2019.

\bibitem{li2022curvature}
H.~Li, J.~Cao, J.~Zhu, Y.~Liu, Q.~Zhu, and G.~Wu, ``Curvature graph neural network,'' \emph{Information Sciences}, vol. 592, pp. 50--66, 2022.

\bibitem{jiang2020drug}
M.~Jiang, Z.~Li, S.~Zhang, S.~Wang, X.~Wang, Q.~Yuan, and Z.~Wei, ``Drug--target affinity prediction using graph neural network and contact maps,'' \emph{RSC Advances}, vol.~10, no.~35, pp. 20\,701--20\,712, 2020.

\bibitem{zhou2020distance}
J.~Zhou, S.~Li, L.~Huang, H.~Xiong, F.~Wang, T.~Xu, H.~Xiong, and D.~Dou, ``Distance-aware molecule graph attention network for drug-target binding affinity prediction,'' \emph{arXiv preprint arXiv:2012.09624}, 2020.

\bibitem{toppingunderstanding}
J.~Topping, F.~Di~Giovanni, B.~P. Chamberlain, X.~Dong, and M.~M. Bronstein, ``Understanding over-squashing and bottlenecks on graphs via curvature,'' in \emph{International Conference on Learning Representations}, 2022.

\bibitem{hamilton1982three}
R.~S. Hamilton, ``Three-manifolds with positive ricci curvature,'' \emph{Journal of Differential Geometry}, vol.~17, no.~2, pp. 255--306, 1982.

\bibitem{colding1997ricci}
T.~H. Colding, ``Ricci curvature and volume convergence,'' \emph{Annals of Mathematics}, vol. 145, no.~3, pp. 477--501, 1997.

\bibitem{samal2018comparative}
A.~Samal, R.~Sreejith, J.~Gu, S.~Liu, E.~Saucan, and J.~Jost, ``Comparative analysis of two discretizations of ricci curvature for complex networks,'' \emph{Scientific Reports}, vol.~8, no.~1, p. 8650, 2018.

\bibitem{ollivier2007ricci}
Y.~Ollivier, ``Ricci curvature of metric spaces,'' \emph{Comptes Rendus Mathematique}, vol. 345, no.~11, pp. 643--646, 2007.

\bibitem{lin2011ricci}
Y.~Lin, L.~Lu, and S.-T. Yau, ``Ricci curvature of graphs,'' \emph{Tohoku Mathematical Journal, Second Series}, vol.~63, no.~4, pp. 605--627, 2011.

\bibitem{ni2018network}
C.-C. Ni, Y.-Y. Lin, J.~Gao, and X.~Gu, ``Network alignment by discrete ollivier-ricci flow,'' in \emph{International Symposium on Graph Drawing and Network Visualization}.\hskip 1em plus 0.5em minus 0.4em\relax Springer, 2018, pp. 447--462.

\bibitem{monge1781memoire}
G.~Monge, ``M{\'e}moire sur la th{\'e}orie des d{\'e}blais et des remblais,'' \emph{Mem. Math. Phys. Acad. Royale Sci.}, pp. 666--704, 1781.

\bibitem{villani2008optimal}
C.~Villani \emph{et~al.}, \emph{Optimal transport: old and new}.\hskip 1em plus 0.5em minus 0.4em\relax Springer, 2008, vol. 338.

\bibitem{ren2024towards}
C.-X. Ren, Y.~Zhai, Y.-W. Luo, and H.~Yan, ``Towards unsupervised domain adaptation via domain-transformer,'' \emph{International Journal of Computer Vision}, vol. 132, no.~12, pp. 6163--6183, 2024.

\bibitem{dunbar2013csar}
J.~B. Dunbar~Jr, R.~D. Smith, K.~L. Damm-Ganamet, A.~Ahmed, E.~X. Esposito, J.~Delproposto, K.~Chinnaswamy, Y.-N. Kang, G.~Kubish, J.~E. Gestwicki \emph{et~al.}, ``Csar data set release 2012: ligands, affinities, complexes, and docking decoys,'' \emph{Journal of Chemical Information and Modeling}, vol.~53, no.~8, pp. 1842--1852, 2013.

\bibitem{somnath2021multi}
V.~R. Somnath, C.~Bunne, and A.~Krause, ``Multi-scale representation learning on proteins,'' \emph{Advances in Neural Information Processing Systems}, vol.~34, pp. 25\,244--25\,255, 2021.

\bibitem{liu2017forging}
Z.~Liu, M.~Su, L.~Han, J.~Liu, Q.~Yang, Y.~Li, and R.~Wang, ``Forging the basis for developing protein--ligand interaction scoring functions,'' \emph{Accounts of Chemical Research}, vol.~50, no.~2, pp. 302--309, 2017.

\bibitem{mysinger2012directory}
M.~M. Mysinger, M.~Carchia, J.~J. Irwin, and B.~K. Shoichet, ``Directory of useful decoys, enhanced (dud-e): better ligands and decoys for better benchmarking,'' \emph{Journal of Medicinal Chemistry}, vol.~55, no.~14, pp. 6582--6594, 2012.

\bibitem{tran2020lit}
V.-K. Tran-Nguyen, C.~Jacquemard, and D.~Rognan, ``Lit-pcba: an unbiased data set for machine learning and virtual screening,'' \emph{Journal of Chemical Information and Modeling}, vol.~60, no.~9, pp. 4263--4273, 2020.

\bibitem{irwin2005zinc}
J.~J. Irwin and B.~K. Shoichet, ``Zinc- a free database of commercially available compounds for virtual screening,'' \emph{Journal of Chemical Information and Modeling}, vol.~45, no.~1, pp. 177--182, 2005.

\bibitem{yang2022mgraphdta}
Z.~Yang, W.~Zhong, L.~Zhao, and C.~Y.-C. Chen, ``Mgraphdta: deep multiscale graph neural network for explainable drug--target binding affinity prediction,'' \emph{Chemical Science}, vol.~13, no.~3, pp. 816--833, 2022.

\bibitem{lim2019predicting}
J.~Lim, S.~Ryu, K.~Park, Y.~J. Choe, J.~Ham, and W.~Y. Kim, ``Predicting drug--target interaction using a novel graph neural network with 3d structure-embedded graph representation,'' \emph{Journal of Chemical Information and Modeling}, vol.~59, no.~9, pp. 3981--3988, 2019.

\bibitem{schutt2018schnet}
K.~T. Sch{\"u}tt, H.~E. Sauceda, P.-J. Kindermans, A.~Tkatchenko, and K.-R. M{\"u}ller, ``Schnet--a deep learning architecture for molecules and materials,'' \emph{The Journal of Chemical Physics}, vol. 148, no.~24, 2018.

\bibitem{satorras2021n}
V.~G. Satorras, E.~Hoogeboom, and M.~Welling, ``E (n) equivariant graph neural networks,'' in \emph{International Conference on Machine Learning}.\hskip 1em plus 0.5em minus 0.4em\relax PMLR, 2021, pp. 9323--9332.

\bibitem{jiang2023metalprognet}
D.~Jiang, Z.~Ye, C.-Y. Hsieh, Z.~Yang, X.~Zhang, Y.~Kang, H.~Du, Z.~Wu, J.~Wang, Y.~Zeng \emph{et~al.}, ``Metalprognet: a structure-based deep graph model for metalloprotein--ligand interaction predictions,'' \emph{Chemical Science}, vol.~14, no.~8, pp. 2054--2069, 2023.

\bibitem{zhang2023ss}
S.~Zhang, Y.~Jin, T.~Liu, Q.~Wang, Z.~Zhang, S.~Zhao, and B.~Shan, ``Ss-gnn: a simple-structured graph neural network for affinity prediction,'' \emph{ACS Omega}, vol.~8, no.~25, pp. 22\,496--22\,507, 2023.

\bibitem{wang2022learning}
L.~Wang, H.~Liu, Y.~Liu, J.~Kurtin, and S.~Ji, ``Learning hierarchical protein representations via complete 3d graph networks,'' \emph{arXiv preprint arXiv:2207.12600}, 2022.

\bibitem{feng2023protein}
S.~Feng, M.~Li, Y.~Jia, W.~Ma, and Y.~Lan, ``Protein-ligand binding representation learning from fine-grained interactions,'' \emph{arXiv preprint arXiv:2311.16160}, 2023.

\bibitem{gainza2020deciphering}
P.~Gainza, F.~Sverrisson, F.~Monti, E.~Rodola, D.~Boscaini, M.~M. Bronstein, and B.~E. Correia, ``Deciphering interaction fingerprints from protein molecular surfaces using geometric deep learning,'' \emph{Nature Methods}, vol.~17, no.~2, pp. 184--192, 2020.

\bibitem{ballester2010machine}
P.~J. Ballester and J.~B. Mitchell, ``A machine learning approach to predicting protein--ligand binding affinity with applications to molecular docking,'' \emph{Bioinformatics}, vol.~26, no.~9, pp. 1169--1175, 2010.

\bibitem{friesner2004glide}
R.~A. Friesner, J.~L. Banks, R.~B. Murphy, T.~A. Halgren, J.~J. Klicic, D.~T. Mainz, M.~P. Repasky, E.~H. Knoll, M.~Shelley, J.~K. Perry \emph{et~al.}, ``Glide: a new approach for rapid, accurate docking and scoring. 1. method and assessment of docking accuracy,'' \emph{Journal of Medicinal Chemistry}, vol.~47, no.~7, pp. 1739--1749, 2004.

\bibitem{marcou2007optimizing}
G.~Marcou and D.~Rognan, ``Optimizing fragment and scaffold docking by use of molecular interaction fingerprints,'' \emph{Journal of Chemical Information and Modeling}, vol.~47, no.~1, pp. 195--207, 2007.

\bibitem{desaphy2013encoding}
J.~Desaphy, E.~Raimbaud, P.~Ducrot, and D.~Rognan, ``Encoding protein--ligand interaction patterns in fingerprints and graphs,'' \emph{Journal of Chemical Information and Modeling}, vol.~53, no.~3, pp. 623--637, 2013.

\bibitem{jain2007surflex}
A.~N. Jain, ``Surflex-dock 2.1: robust performance from ligand energetic modeling, ring flexibility, and knowledge-based search,'' \emph{Journal of Computer-Aided Molecular Design}, vol.~21, no.~5, pp. 281--306, 2007.

\bibitem{wojcikowski2017performance}
M.~W{\'o}jcikowski, P.~J. Ballester, and P.~Siedlecki, ``Performance of machine-learning scoring functions in structure-based virtual screening,'' \emph{Scientific Reports}, vol.~7, no.~1, p. 46710, 2017.

\end{thebibliography}

\end{document}